\documentclass[10pt,journal,compsoc]{IEEEtran}
\usepackage{amssymb}
\usepackage{bbm}
\usepackage{amsfonts}
\usepackage{multirow}
\usepackage{booktabs}
\usepackage{color}
\usepackage{makecell}
\usepackage{easyReview}
\usepackage{threeparttable}
\usepackage{graphicx}
\usepackage{colortbl}
\usepackage{nicematrix}
\usepackage{tabularx}
\usepackage{xcolor}
\usepackage{ragged2e}
\DeclareUnicodeCharacter{FB01}{fi}
\ifCLASSOPTIONcompsoc
  \usepackage[nocompress]{cite}
\else
  \usepackage{cite}
\fi

\usepackage{amsmath}
\newcommand{\newrevise}[1]{\textcolor{black}{#1}}
\title{Spatial-Temporal Cross-View Contrastive Pre-training for Check-in Sequence Representation Learning}

\author{Letian~Gong,
    Huaiyu~Wan,
    Shengnan~Guo*,
    Xiucheng~Li,
    Yan~Lin\\
    Erwen~Zheng,
    Tianyi~Wang,
    Zeyu~Zhou,
    Youfang~Lin
	\IEEEcompsocitemizethanks{\IEEEcompsocthanksitem L. Gong, H. Wan, S. Guo, Y. Lin, E. Zheng, T. Wang, Z.Zhou, Y. Lin are with the Key Laboratory of Big Data \& Artificial Intelligence in Transportation, Ministry of Education, Beijing Jiaotong University, Beijing 100044, China, and the Key Laboratory of Intelligent Passenger Service of Civil Aviation, CAAC, Beijing, 101318, China.\protect\\
           E-mail: gonglt@bjtu.edu.cn; hywan@bjtu.edu.cn; guoshn@bjtu.edu.cn; ylincs@bjtu.edu.cn; zhengerwen@bjtu.edu.cn; wangtianyi@bjtu.edu.cn; zhouzeyu@bjtu.edu.cn; yflin@bjtu.edu.cn.
     \IEEEcompsocthanksitem X. Li is with the School of Computer Science and Technology, Harbin Institute of Technology, Shenzhen,  518055, China. \protect\\
		  E-mail: lixiucheng@hit.edu.cn.}
	\thanks{~\protect\\(Corresponding author: Shengnan Guo.)}}

\newtheorem{definition}{Definition}	
\newcommand{\paratitle}[1]{\vspace{1ex}\noindent \textbf{#1}}
\makeatletter
\def\endthebibliography{%
	\def\@noitemerr{\@latex@warning{Empty `thebibliography' environment}}%
	\endlist
}
\makeatother
\begin{document}

\IEEEtitleabstractindextext{%
\begin{abstract}
\justifying
The rapid growth of location-based services (LBS) has yielded massive amounts of data on human mobility. Effectively extracting meaningful representations for user-generated check-in sequences is pivotal for facilitating various downstream services. However, the user-generated check-in data are simultaneously influenced by the surrounding objective circumstances and the user's subjective intention. Specifically, the temporal uncertainty and spatial diversity exhibited in check-in data make it difficult to capture the macroscopic spatial-temporal patterns of users and to understand the semantics of user mobility activities. Furthermore, the distinct characteristics of the temporal and spatial information in check-in sequences call for an effective fusion method to incorporate these two types of information. In this paper, we propose a novel Spatial-Temporal Cross-view  Contrastive Representation (STCCR) framework for check-in sequence representation learning. Specifically, STCCR addresses the above challenges by employing self-supervision from "spatial topic" and "temporal intention" views, facilitating effective fusion of spatial and temporal information at the semantic level. Besides, STCCR leverages contrastive clustering to uncover users' shared spatial topics from diverse mobility activities, while employing angular momentum contrast to mitigate the impact of temporal uncertainty and noise. We extensively evaluate STCCR on three real-world datasets and demonstrate its superior performance across three downstream tasks.
\end{abstract}

}

\maketitle
\IEEEdisplaynontitleabstractindextext
\IEEEpeerreviewmaketitle

\IEEEraisesectionheading{
\section{Introduction}\label{sec:introduction}
}

\IEEEPARstart {L}{ocation-based} services (LBS), such as Gowalla, Weeplace, and Yelp, have experienced significant development over the past decade. These platforms enable users to share and discover location information and surrounding services, resulting in the accumulation of extensive data on human mobility behavior, \emph{e.g.} check-in sequences at points of interest (POIs). This offers prospects for analyzing and comprehending human mobility patterns for various practical applications, such as predicting the next check-in location or time for personalized recommendations, linking trajectories to users, and detecting abnormal trajectories for safety control purposes \emph{etc.}

Learning accurate and universal representations for check-in sequences is a crucial task in human mobility data mining. However, the existing excellent end-to-end models for check-in sequences modeling, such as those designed for location prediction~\cite{DeepMove, LSTPM,SERM}, time prediction~\cite{IFLTPP}, and trajectory user link~\cite{DeepTUL, TULER,TULVAE}, often struggle to learn generalized representations for check-in sequences and fail to comprehensively describe the spatial-temporal patterns and high-level semantics of human mobility, since the supervision signals of these models usually rely on limited single-type labels. Therefore, the learned representations are task-specific and poorly generalized. To facilitate the generalization ability for the check-in sequence's representations, pre-training check-in sequence representation via self-supervised learning has been widely studied and proven to be an effective way to fully exploit massive unlabeled check-in data to boost the performance of the downstream tasks. 

Representation learning is always one of the hot research topics in deep learning. And recently, contrastive pre-training with self-supervised signals~\cite{self-supervised} has emerged as the most effective approach for sequence modeling.  
In particular, some representative works~\cite{ReMVC,SML,CACSR} in the spatial-temporal data mining (STDM) domain have proven their effectiveness in learning the representations of check-in sequences. 
However, the unique spatial and temporal characteristics of check-in sequences raise challenges for these contrastive pre-training based models, meanwhile weakening their ability to capture the macroscopic spatial-temporal mobility patterns and to understand the high-level semantics of user mobility activities.
Specifically, we identify three key challenges: 
%


\begin{figure}[htb]
	\centering
	\includegraphics[width=1.00\columnwidth]{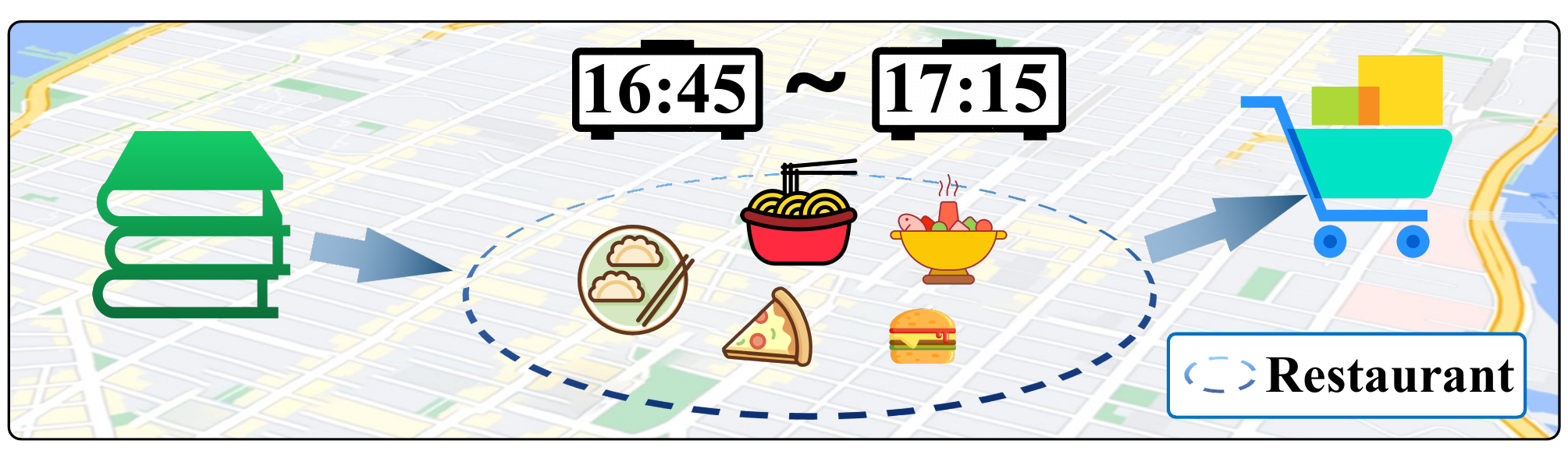}
	\vspace{-2em}
    \caption{shows the temporal uncertainty is influenced by the subjective intention and objective factors. Users' arrival times tend to be in a range of intervals rather than a precise planned time.}
	\label{intro1}
 \vspace{-0em}
\end{figure}


\newrevise{(1) \textbf{Temporal uncertainty}: Understanding the \textit{temporal intention} of users' ability from the check-in sequence with uncertain temporal information is challenging. As shown in Fig.~\ref{intro1}, based on the user’s historical sequence and the user’s historical spatial-temporal behavior patterns, the user is most likely to go for dinner next, with the strongest temporal intention being at 17:00. However, the exact time of arrival is simultaneously influenced by his/her subjective decisions such as the today's choice of restaurant, and the surrounding objective factors such as traffic and weather. This leads to the temporal uncertainty, \emph{i.e.}, the user may arrive at the restaurant for dinner around 16:45 to 17:15, rather than the precise time of 17:00.} 
Besides, unexpected bugs on service platforms may also bring noise to the recorded check-in time. 
The uncertainty and noise make it difficult to extract the user's temporal intention from the raw temporal context of the check-in sequences. 
Most existing check-in sequence representation learning methods capture the temporal patterns by embedding the precise time of check-ins, overlooking the inherent uncertainty presented in the temporal dimension of check-in sequences. 
Therefore, it is difficult for these methods to explore the periodic patterns and to understand the temporal intention of the users' mobility relying on noisy time. 

\begin{figure}[htb]
	\centering
	\includegraphics[width=1.00\columnwidth]{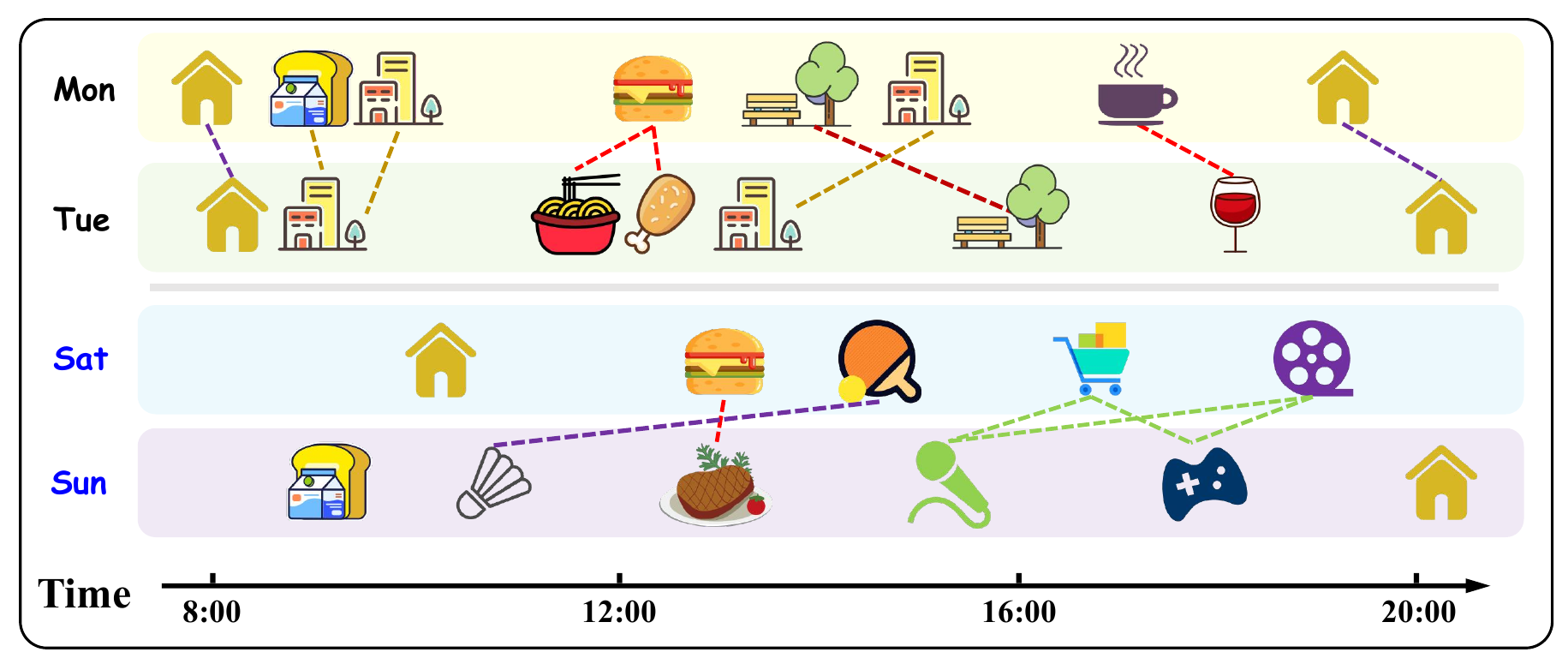}\vspace{-0em}
	\caption{shows the check-in sequences of a user on working days and weekends, and distinct icons indicate different POIs. Although the check-in POIs on the four days differ, the semantics reflected by the check-in sequences on the two working days (or the two weekends) are similar.}
	\label{intro2}
\end{figure}

(2) \textbf{Spatial diversity}: Users' mobility has a high degree of diversity over POIs. Fig.~\ref{intro2} shows the check-in sequences of a user on consecutive working days and weekends. 
It can be seen that there are obvious differences between the user's \textit{spatial topic} of working days and weekends. Specifically, the mobility on working days is mainly centered around office-related POIs, while that on weekends is mainly centered around leisure and entertainment POIs. 
Meanwhile, although the semantics reflected by check-in sequences on the two working days (or the two weekends) are similar, the specific POIs on the two working days (or the two weekends) may be diverse and rarely repeated. We term this phenomenon spatial diversity, which prevents us from effectively capturing the shared mobility patterns between check-in sequences with similar semantics but different POIs. 
Existing prevalent check-in sequence representation learning methods often adopt the most straightforward word embedding strategy borrowed from the natural language processing (NLP) domain to represent the POIs, \emph{i.e.}, learning embeddings for POIs and retrieving them using indices. 
That is, these methods treat each POI individually and fail to capture the shared semantics reflected by the POI sequences, resulting in the inability of these models to capture the high-level spatial-temporal semantics of users' mobility. 

(3) As discussed above, the raw check-in sequence is discretized and diverse w.r.t. the spatial information, but continuous and uncertain in w.r.t. the temporal information. Consequently, the distinct properties of spatial and temporal information make it challenging to effectively fuse them together. %
%
%
Some existing models adopt a "fuse prior to modeling" approach to model the temporal and spatial information of check-in sequences. 
However, these models encounter difficulties in simultaneously capturing spatial and temporal semantics within a unified encoder, since they ignore the difference between temporal and spatial information.
Furthermore, raw data are at a fine-grained level and barely contain worthwhile semantic information. Besides, there are some disturbances in the raw data, such as temporal noise and spatial diversity. Therefore, combining the fine-grained level properties of temporal and spatial at an early phase would disrupt the performance of the model in relation to each other~\cite{MFLVC}. 
Other models explore using separate encoders to learn temporal and spatial representations before fusion. However, the learned temporal and spatial representations reside in separate spaces and lack alignment~\cite{ALBEF}. They usually use direct collocation or gating mechanisms, ignoring the relevance between spatial topic and temporal intention.

To overcome the aforementioned limitations, we propose the \underline{S}patial-\underline{T}emporal \underline{C}ross-view \underline{C}ontrastive \underline{R}epresentation (STCCR), a pre-training framework for learning the representations of check-in sequences. 
Our method aims to achieve an effective fusion manner that preserves potential spatial-temporal cross-view correlations at the macroscopic semantic level. 
By treating one view as the reference, our cross-view contrastive strategy facilitates spatial-temporal information interaction, generating numerous high-quality self-supervisions.
Additionally, STCCR learns spatial topics from all check-in sequences using contrastive clustering, capturing topics by exploring the shared mobility pattern from diverse human behavior.
In the temporal intention view, we employ an angular margin manner. This leverages angular margin self-supervised signals to mitigate the effects of temporal uncertainty and noise on the angular margin. 
In summary, our contributions are as follows.

 \begin{itemize}
            \item We propose a novel spatial-temporal cross-view contrastive framework for check-in sequence representation learning from the spatial topic and temporal intention views. To the best of our knowledge, this is the first study to leverage a cross-view contrastive manner to explore the spatial-temporal correlation of human mobility at the macroscopic semantic level.

            \item  We propose an angular margin contrast-based method to exploit the inherent uncertainty of the time information in check-ins. By adding a soft interval to the contrast learning training, the temporal noise information can be filtered so that the model can effectively capture the user's temporal intention.
            
             \item We perform contrastive clustering in the spatial dimension. To address the diversity of POIs, we explore shared spatial topics by clustering high-level semantic information from check-in sequences.
             
            \item We evaluate STCCR on three real-world datasets for three downstream tasks. The experimental results prove the superiority and versatility of our model. 
          
 \end{itemize}  

\section{Related work}
\subsection{Mobility Data Mining}
Location-based services have given rise to a new and promising research topic known as mobility data mining, which has led to the emergence of three significant tasks that contribute to enhancing the quality of services: next location prediction (LP), next time prediction (TP), and trajectory user link (TUL). 
Recent studies have confirmed that deep learning techniques, specifically recurrent neural networks (RNNs) and attention mechanisms, are highly effective in capturing sequential and periodic patterns of human mobility. By combining deep learning techniques, researchers have made significant advancements in capturing both the sequential and periodic patterns of human mobility. The core of these models is the modeling of check-in sequences, which leads to improved accuracy in location prediction and trajectory analysis.

LP aims to anticipate a user's future location based on their historical movement. Several notable models have emerged as prominent approaches in LP. DeepMove~\cite{DeepMove} leverages RNNs and attention mechanisms to capture the spatial-temporal intentions in users' location data and predict their next destination. STAN~\cite{STAN} introduces a spatial-temporal attention network that incorporates spatial and temporal contexts for accurate prediction. LSTPM~\cite{LSTPM} focuses on long and short-term patterns in user trajectory using an attention-based LSTM~\cite{LSTM} model. SERM~\cite{SERM} utilizes an encoder-decoder architecture with a spatial-temporal residual network to capture user preferences and predict future locations. PLSPL~\cite{PLSPL} trains two LSTM models for location and category based sequence to capture the user's preference. LightMove~\cite{LightMove} designs neural ordinary differential equations to enhance robustness against sparse or incorrect inputs. HMT-GRN~\cite{HMT-GRN} alleviates the data sparsity problem by learning different User-Region matrices of lower sparsities in a multitask setting. Graph-Flashback~\cite{Graph-Flashback} constructs a spatial-temporal knowledge graph to enhance the representation of POIs. GETNext~\cite{GETNext} introduces a user-agnostic global trajectory flow map as a means to leverage the abundant collaborative signals.

TUL is a significant task that focuses on establishing connections between different trajectories, facilitating the analysis of user movement patterns, and uncovering valuable insights about their behavior. Notable models have been specifically developed to address the challenge of predicting trajectory links. TULER~\cite{TULER} takes advantage of advanced algorithms to establish links between trajectories, allowing for a comprehensive understanding of user movement patterns. DeepTUL~\cite{DeepTUL} utilizes deep learning techniques to extract representations from trajectory data and facilitate the prediction of trajectory links. S2TUL~\cite{S2TUL} utilizes graph convolutional networks and sequential neural networks to capture trajectory relationships and intra-trajectory information. GNNTUL~\cite{GNNTUL} employs graph neural networks for human mobility and associates the traces with users on social networks.

TP focuses on estimating the time at which a user is likely to visit their next location. To accomplish this, it is common practice to use intensity functions to represent the rate or density of event occurrences, various models have been developed to model the intensity function and make accurate time predictions effectively. Modeling the intensity function using RNNs or attention mechanisms is a common approach for predicting the occurrence of events. RMTPP~\cite{RMTPP} utilizes RNNs to model the intensity function. SAHP~\cite{SAHP} combines the Hawkes process with self-attention mechanisms to capture the temporal dependencies and spatial influences in event sequences. THP~\cite{THP} combines the Hawkes process with transformer-based architectures to capture temporal dependencies in event sequences. NSTPP~\cite{NSTPP} utilizes neural ODEs to model discrete events in continuous time and space, enabling the learning of complex distributions in spatial and temporal domains. IMTPP~\cite{IMTPP} models the generative processes of observed and missing events and utilizes unsupervised modeling and inference methods for time prediction. DSTPP~\cite{DSTPP} purposes a novel parameterization framework that uses diffusion models to learn complex joint distributions.

It is important to note that these end-to-end supervised methods designed for specific tasks are not universal. These models do not have a good grasp of the macroscopic semantics of check-in sequences. Thus, learning the universal representation of check-in sequences to improve the model's ability and understand high-level semantics is critical.

\subsection{Pretraining and Contrastive Learning}
The essence of mobility mining tasks lies in learning the representation of check-in sequences. Numerous studies have demonstrated the effectiveness of employing the pre-training paradigm to achieve check-in sequence representation learning. For instance, TULVAE~\cite{TULVAE} and MoveSim~\cite{MoveSim} utilize Variational Auto-Encoder and Generative Adversarial Network, respectively, to capture the movement patterns of check-in sequences through pre-training. TALE~\cite{TALE} proposes a pre-training representation scheme for trajectory point location embedding that incorporates temporal semantics, which effectively improves the performance of next location prediction and location traffic prediction. CTLE~\cite{CTLE} proposes a location pre-training representation model that incorporates domain features, which dynamically generates feature representations of the domain environment of the target location so that the model can better capture the macroscopic higher-order semantic information in the latitude and longitude.

As a kind of advanced SSL technology, contrastive learning-based pre-training techniques have demonstrated great potential in the field of Natural Language Processing (NLP). It utilizes self-supervised training by comparing positive and negative pairs generated through data augmentation., contrastive pre-training models employ a variety of data augmentation strategies. For example, SimCSE~\cite{SimCSE} utilizes dropout operations for data augmentation. ConSERT~\cite{ConSERT} disrupts, slices, and deletes representations in the hidden space. VaSCL~\cite{VaSCL} enhances the discriminative power by introducing challenging negative samples. CLAPS~\cite{CLAPS} introduces adversarial perturbations to generate indistinguishable augmented samples, thus significantly improving the robustness and discrimination ability.

In the domain of mobility mining, the first model to adopt contrastive learning is SML~\cite{SML}, which applies commonly used data augmentation strategies such as cropping or replacement to check-in sequences. ReMVC~\cite{ReMVC} learns distinct region embeddings and constraints embedding parameters while transferring knowledge across multiple views. DRAN~\cite{DRAN} exploits disentangled representations to capture distinct aspects and corresponding influences for a more precise representation of POIs. CACSR~\cite{CACSR} proposes a contrastive pre-training model for learning check-in sequence representations with adversarial perturbations. However, these models are not designed separately for spatial and temporal properties and do not specifically consider spatial-temporal information fusion.

In summary, learning about check-in sequence representation is crucial for mobility mining tasks. Pre-training methods, especially contrastive learning, have demonstrated effectiveness in capturing underlying patterns. It is essential to develop tailored techniques that can effectively extract human spatial-temporal patterns of check-in sequences to improve the performance of contrastive pre-training models in this domain.

\begin{figure*}[htbp]
    \centering
    \includegraphics[width=1\textwidth]{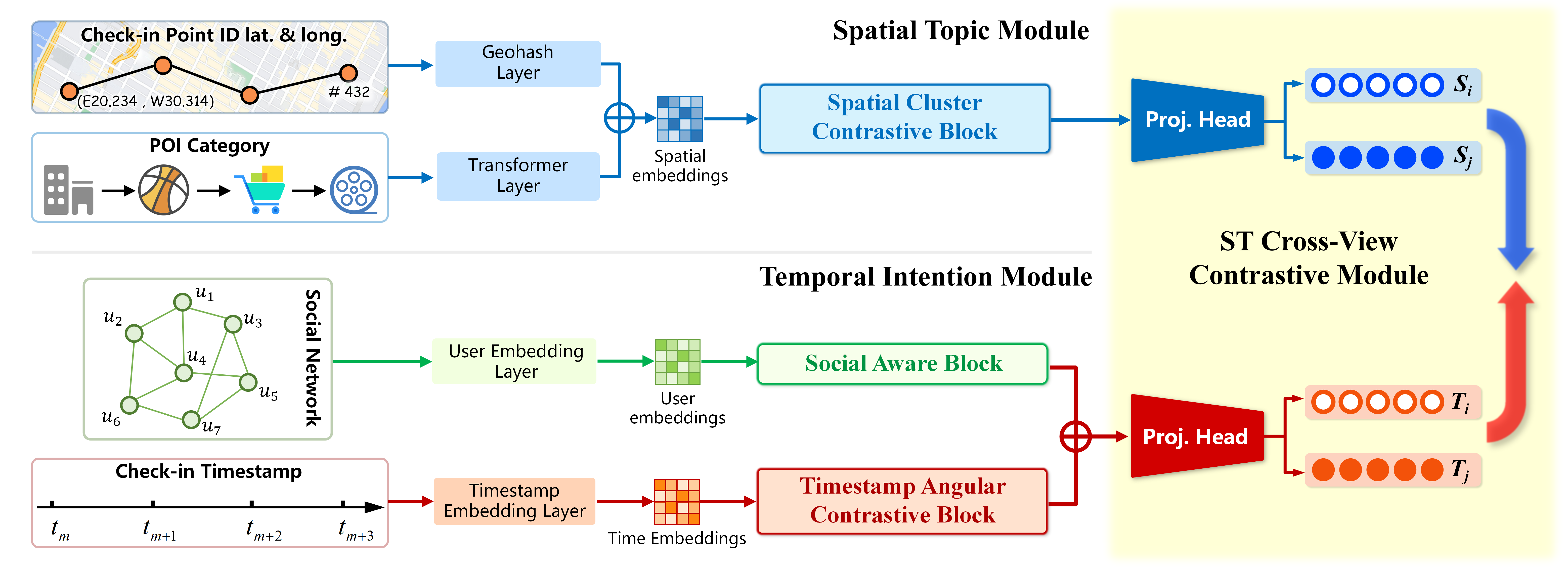}
    \caption{The model architecture of STCCR. (a)The Spatial Topic Module employs contrast clustering to encode the POIs' id, latitude, longitude, and category. (b)The Temporal Intention Module combines user and time information to obtain the temporal intention patterns of users. (c)The ST Cross-View Contrastive Module aligns spatial and temporal information into a unified semantic space using project heads, facilitating the integration of spatial-temporal information at the macroscopic semantic level.}
    \label{TKDE_Gong_overall}
\end{figure*}

\section{Preliminaries}
\subsection{Definitions}
\begin{definition}\textbf{POI Visiting Record.} 
In location-based services datasets, a user's visit to a certain place is represented by a POI visiting record $r = (u, l, t)$, where $u$ represents the user, $l$ indicates the visited location, and $t$ denotes the timestamp of the visit.
The location $l$ is represented by $(lid,lon,lat,c)$, comprising $lid$ as a POI index or a grid index, and accurate longitude $lon$ and latitude $lat$. $c$ denotes the category of the visited location (\emph{e.g.}, hospital or restaurant). 
\end{definition}

\begin{definition}\textbf{Check-in Sequence.}
The movement of a user during a specific period can be represented by a list of sequential POI visiting records, which we refer to as a check-in sequence. 
We denote a check-in sequence as $\mathcal{T}=<r_1, r_2, \cdots, r_s>$, where the POI visit records are ordered by their visited time, and $s$ is the length of the sequence.
\end{definition}

\subsection{Problem Statement} 
\textit{Pre-training Representation for check-in Sequence.} 
The goal of this paper is to pre-train a parameterized encoder $G$ capable of generating a contextual representation for a given check-in sequence $\mathcal{T}$, denoted as $G(\mathcal{T})$. 
Specifically, the encoder $G$ is first trained within a spatial-temporal cross-view framework using a contrastive manner, without task-specific objectives. Then, it can be applied to various downstream tasks, such as next Location Prediction (LP), Trajectory User Link (TUL), and Time Prediction (TP), among others. 
We expect that the parameterized encoder $G$ can be widely used to enhance the performance of these downstream tasks.

\section{spatial-temporal Cross-view Contrastive Framework}
As illustrated in Fig.~\ref{TKDE_Gong_overall}, we propose a \textbf{S}patial \textbf{T}emporal \textbf{C}ross-view \textbf{C}ontrastive \textbf{R}epresentation (STCCR) model that leverages self-supervision to capture high-level semantics, i.e., the spatial topic and temporal intention of check-in sequences separately and then fuse them at a macroscopic level. 
To extract the shared spatial topic of the check-in sequence, we introduce the \textbf{S}patial \textbf{T}opic \textbf{M}odule (STM). 
This module employs contrastive clustering to encode the spatial information of check-in sequences, forcing check-in sequences with similar spatial topics to have similar representations.
Additionally, we combine time and user information in the \textbf{T}emporal \textbf{I}ntention \textbf{M}odule (TIM) during pre-training to extract the temporal intention of users. Specifically, we adopt a contrastive learning scheme with an angular margin to model noisy temporal information.
Finally, the ST Cross-View Contrastive Module aligns the high-level spatial and temporal semantics into a unified semantic space using project heads, facilitating the integration of spatial-temporal information at the macroscopic semantic level.
Next, we provide a detailed explanation of our proposed model in the following sections.

\subsection{Spatial Topic Module}
The Spatial Topic Module comprises a geohash layer, a transformer layer, and a spatial cluster contrastive block. The geohash layer and transformer layer work together to embed geographical location information into the embedding space. The contrastive spatial cluster block leverages contrastive clustering to capture spatial topics of users' mobility.

\subsubsection{Geographical Location Information Encoding}
The key advantage of Geohash~\footnote{https://geohash.co} encoding is its ability to convert geographic coordinates into a string of characters, enabling efficient storage, retrieval, and analysis of location-based data. Geohash represents latitude and longitude information in the following three steps. 
First, the latitude and longitude are converted into two binary sequences, $e_{lat}$ and $e_{lon}$. These sequences are obtained by recursively dividing the latitude and longitude ranges. For the latitude value, the range $(-90^{\circ}, 90^{\circ})$ is divided into two sub-ranges: $(-90^{\circ}, 0)$ and $(0, 90^{\circ})$. If the latitude falls within the lower sub-range, a '0' is appended to $e_{lat}$, otherwise a '1' is appended. The same process is applied to $e_{lon}$ using the initial range $(-180^{\circ}, 180^{\circ})$. 
Next, the even bits of $e_{geo}$ are set to $e_{lat}$ and the odd bits are set to $e_{lon}$ to create the concatenated binary sequence $e_{geo}$, where $i = \{0,1,2,\cdots,15\}$. 
\begin{equation}
    \begin{aligned}
    e_{(geo, 2i)} & = e_{(lat,i)} \\
    e_{(geo, 2i+1)} & =  e_{(lon,i)}  
    \end{aligned}
\end{equation}
Finally, $e_{geo}$ is converted to Base32 encoding to produce the geohash representation.

\subsubsection{POI Category representation}
To represent the category description of a POI, we treat the description as words and directly utilize a public pre-trained BERT model\footnote{https://huggingface.co} for sequence representation. \newrevise{We use a variant of BERT} in which the final output of the [CLS] token is selected as the representation of the description. The representation of a POI description is denoted as $e_{cat}$.

\subsubsection{Spatial Cluster Contrastive Block}
To gain a more comprehensive understanding of user mobility patterns, we propose a spatial cluster contrastive block to capture the underlying shared spatial topics of users' mobility.
As discussed in Section~\ref{sec:introduction}, we can find a great deal of diversity among the POIs of check-in sequences. Thus, treating each individual check-in sequence separately without considering their common mobility patterns makes it hard to share statistical strength across sequences.
We find that users tend to exhibit movement patterns centered on specific spatial topics during different periods. For example, users tend to move around work areas, dining areas, and residential areas during the working days, while focusing on leisure activities such as travel areas, shopping centers, and dining areas during the weekends. Therefore, extracting spatial topics from diverse check-in sequences is crucial to effectively learning the shared mobility patterns of users' movement.

That is, spatial topics refer to the check-in sequences that are generated over different locations but have similar, relative, and shared patterns in terms of spatial movement. To explore the shared spatial topics among sequences, we introduce the "clustering consistency" and "reweighted contrastive" strategies into our model. 
To represent different shared spatial topics, we define a prototype $\boldsymbol{C}$ which is a set of $k$ cluster centers $\boldsymbol{C}=\{\boldsymbol{c}_{1}, \cdots, \boldsymbol{c}_{K}\}$. 
Meanwhile, we assume that check-in sequences with the same spatial topics fall into a similar semantic space. 
We use a Bi-GRU as the spatial encoder to encode the spatial information of check-in sequences combined by $e_{geo}$ and $e_{cat}$. Given a representation of the check-in sequence through the spatial encoder $\boldsymbol{z}_{s}^n$ as the anchor, we use the dropout augmentation manner as SimCSE~\cite{SimCSE} to obtain its augmentation $\boldsymbol{z}_{s}^m$. We calculate each prototype assignment $\boldsymbol{q}_{i}$ by assessing the similarity of the representation to the prototype as follows:
\begin{equation}
    \quad \boldsymbol{q}_{i}^{(k)}  =\frac{\exp \frac{\boldsymbol{z}_{s}^{j\top} \boldsymbol{c}_{k}}{\tau}}{\sum_{k^{\prime} \neq k} \exp \frac{\boldsymbol{z}_{s}^{j\top} \boldsymbol{c}_{k^{\prime}}}{\tau}},
\end{equation}
where each $\boldsymbol{q}_{i} = [\boldsymbol{q}_{i}^{(1)}, \boldsymbol{q}_{i}^{(2)}, \cdots , \boldsymbol{q}_{i}^{(k)}, \cdots, \boldsymbol{q}_{i}^{(K)}]$, $i=\{1,2,\cdots,N\}$, $N$ is the total number of check-in sequences, $\tau$ is a temperature parameter. To ensure the consistency of class attribution between the anchor and augmented sample, we define the clustering consistency loss function as:
\begin{equation}
\mathcal{L}_{\text{C}}\left(\boldsymbol{z}_{s}^n, \boldsymbol{z}_{s}^m\right)=\ell\left(\boldsymbol{z}_{s}^m, \boldsymbol{q}_{n}\right)+\ell\left(\boldsymbol{z}_{s}^n, \boldsymbol{q}_{m}\right),
\label{eq1}
\end{equation}
where $\ell(\boldsymbol{z}_s, \boldsymbol{q})$ measures the fit between representation $\boldsymbol{z}_s$ and assignment $\boldsymbol{q}$. We compare the representations $\boldsymbol{z}_{s}^n$ and $\boldsymbol{z}_{s}^m$ using their prototype assignments $\boldsymbol{q}_{n}$ and $\boldsymbol{q}_{m}$. Each term in Eq.~\ref{eq1} represents the cross-entropy loss between $\boldsymbol{q}$ and the probability obtained by taking a softmax of the dot products of $\boldsymbol{z}_{s}$ and all columns in $\boldsymbol{C}$, i.e.,
\begin{equation}
    \ell\left(\boldsymbol{z}_{s}^m, \boldsymbol{q}_{n}\right)  =-\sum_{k} \boldsymbol{q}_{n}^{(k)} \log \boldsymbol{q}_{m}^{(k)}
\end{equation}

Consistency loss makes the anchor and its corresponding sample belong to a similar assignment as much as possible. But this may lead to a plain solution (i.e., the model assigns all samples inside one cluster). To avoid this, we propose a reweighting strategy that assigns larger weights to meaningful negative samples with a moderate prototype distance to the anchor, and smaller weights to negative samples that are easily distinguished. This assigns the samples in the batch and queue to the $K$ classes according to $\boldsymbol{q}$ while satisfying the inter-cluster balance constraint. This makes samples that have similar prototype assignments grouped together as much as possible and avoids the plain-solution problem. We define the reweighting strategy denoted $\mathcal{L}_{R}$ as:
\begin{equation}
    \begin{aligned}
        \mathcal{L}_{R}=-\sum_{n=1}^{N} \quad\log\frac{\phi \left(n,m\right)}{\phi \left(n,m\right)+M_{n} \sum_{j \in S} \boldsymbol{w}_{nj} \phi \left(n,j\right)},
    \end{aligned}
\end{equation}
where $\phi \left(n,m\right)=\exp \left(\boldsymbol{z}^{n\top}_{s}  \boldsymbol{z}_{s}^{m}/ \tau\right)$, $ \boldsymbol{w}_{nj} $ is the weight of negative pairs $\left(\boldsymbol{z}^{n}_{s}, \boldsymbol{z}^{j}_{s}\right)$. $M_{n}=2\beta/(\sum_{j\in S} \boldsymbol{w}_{n j})$ is the normalization factor, $\beta$ is the number of the set $S$. $S=\{j | \boldsymbol{c}^j\neq \boldsymbol{c}^n\}$, where $\boldsymbol{c}^j$ and $\boldsymbol{c}^n$ are the most probable prototypes of the check-in sequence $\boldsymbol{z}_{s}^{j}$ and $\boldsymbol{z}_{s}^{n}$, respectively. 

\begin{figure}[htb]
	\centering
	\includegraphics[width=1\columnwidth]{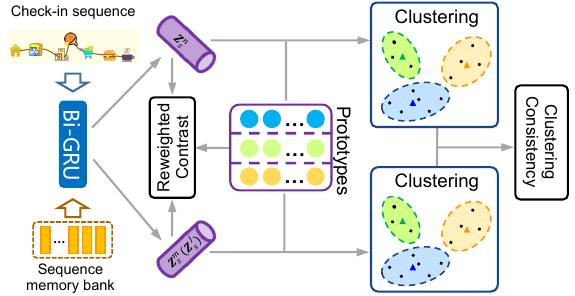}
	\caption{Spatial cluster contrastive block. We capture spatial topic of user activity through reweighted contrast and cluster consistency manners. In order to improve the clustering effect, we maintain a queue of historical sequences that participate in the computation of the current batch.}
	\label{prototype}
\end{figure}

We utilize the cosine distance to measure the distance between two assignments $\boldsymbol{q}_{n}$ and $ \boldsymbol{q}_{j} $ as: $\mathcal{D}\left(\boldsymbol{q}_{n}, \boldsymbol{q}_{j}\right)=1-\left(\boldsymbol{q}_{n} \cdot \boldsymbol{q}_{j}\right) /\left(\left\|\boldsymbol{q}_{n}\right\|_{2}\left\|\boldsymbol{q}_{j}\right\|_{2}\right)$. Then, we define the weight based on the above assignment distance with the format of the Gaussian function as:
\begin{equation}
    w_{n j}=\exp \left\{-\frac{\left[\mathcal{D}\left(\boldsymbol{q}_{n}, \boldsymbol{q}_{j}\right)-\mu_{n}\right]^{2}}{2 \sigma_{n}^{2}}\right\},
\end{equation}
where $\mu_{n}$ and $\sigma_{n} $ are the mean and standard deviation of $\mathcal{D}\left(\boldsymbol{q}_{n}, \boldsymbol{q}_{j}\right) $ for anchor $\boldsymbol{z}^{n}_{s} $, respectively. In this way, selected negative samples can enjoy desirable semantic differences from the anchor, and those similar ones are "masked" out in the objective.

Since different clusters represent distinct underlying semantics, such a sampling strategy can ensure a distinguishable semantic difference between the anchor and its negatives. 
The final training objective is the combination of $\mathcal{L}_{R} $ and $\mathcal{L}_{C}$ to jointly optimise the spatial topic, formulated as:
\begin{equation}
    \mathcal{L}_{Spatial}=\eta\mathcal{L}_{C}+\mathcal{L}_{R}
\end{equation}
where the constant $\eta$ balances the clustering consistency loss $\mathcal{L}_{C}$  and the reweighted contrastive loss $\mathcal{L}_{R}$. This loss function is jointly minimized concerning the prototype $\boldsymbol{C}$ and the parameters $\theta $ of the spatial encoder used to produce the spatial representation $\boldsymbol{z}_{s}$.

\subsection{Temporal Intention Module}
The Temporal Intention Module aims at analyzing users' temporal intentions. It includes a timestamp angular contrastive block and a social aware block. They leverage the angular margin to mitigate the effects of temporal uncertainty and noise. 
\subsubsection{Timestamp Embedding}
The timestamp embedding layers convert the original temporal features into dense vectors. Specifically, we first discretize each timestamp $t$ into hourly intervals and then represent it as a ${T}$-dimensional one-hot vector ($T$ = 48). To distinguish weekends from weekdays, we treat weekends as an additional 24 hours. Then we learn the embedding for each time interval, denoted by $E_t \in \mathbb{R}^{T \times d_{t}}$.
\subsubsection{Temporal Angular Contrastive Block}
This block leverages the angular margin scheme to enable contrastive learning to mitigate the effects of temporal noise and extract users' temporal intention. To model the positive and negative pairwise relations between sequences, we first generate sequence representations and group them into positive and negative pairs. We then use these pairs as input to a training objective for optimization.

\begin{figure}[htb]
	\centering
	\includegraphics[width=1\columnwidth]{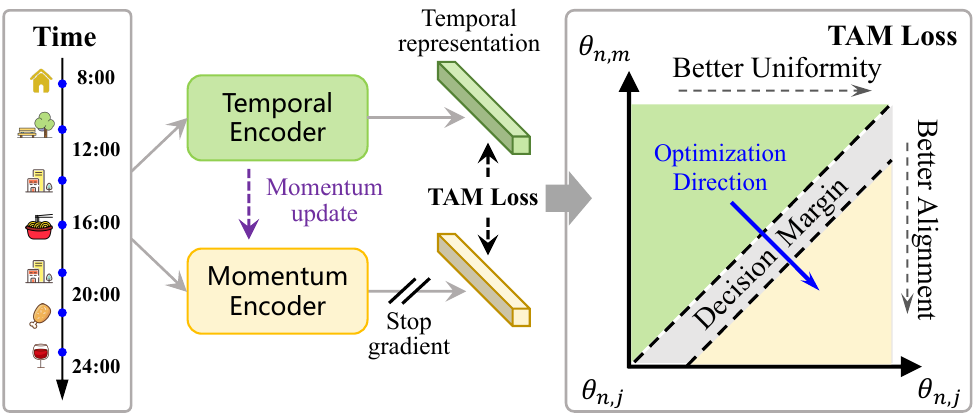}
	\caption{Angular Margin. Better Uniformity refers to the ability of a model to learn shared representations among similar samples, resulting in improved consistency within the feature space. Better Alignment signifies the model's capability to map different views or variations of the same sample to nearby positions in the feature space, achieving enhanced alignment.}
	\label{Angular}
\end{figure}

To generate temporal representations from check-in sequences in the temporal dimension, we employ two Bi-GRU encoders, denoted as $\mathcal{M}_t$ and $\mathcal{M}_t^{\prime}$. Similarly to the approach in ESimCSE~\cite{EsimCSE}, we use momentum contrast as the data augmentation method. In particular, we use the momentum-updated encoder to encode the enqueued sequence representation. Formally, denoting the parameters of the encoder $\mathcal{M}_t$ as $\theta_{t}$ and those of the momentum-updated encoder $\mathcal{M}_t^{\prime}$ as $\theta_{t}^{\prime}$, we update $\theta_{t}^{\prime} $ in the following way:
\begin{equation}
    \theta_{t}^{\prime} \leftarrow \eta \theta_{t}^{\prime} +(1-\eta) \theta_{t},
\end{equation}
where $\eta \in [0,1) $ is a momentum coefficient parameter. Note that only the parameters $\theta_{t}$ are updated by back-propagation. To generate temporal representations, we introduce a new set of parameters denoted as $\theta_{t}^{\prime}$, which are updated using momentum to ensure a smoother evolution than $\theta_{t}$. We obtain two different temporal representations, denoted as the anchor $\boldsymbol{z}_{t}^{n}$ and the augmentation $\boldsymbol{z}_{t}^{m}$, by passing it through the models $\mathcal{M}_t$ and $\mathcal{M}_t^{\prime}$, respectively. These representations share the same semantics and form a positive pair, while negative pairs are obtained by comparing representations from different samples in the same batch. The most widely adopted training objective is the NT-Xent loss, which is formulated as follows:
\begin{equation}
    \mathcal{L}_{\text {NT-Xent}}=-\log \frac{e^{\operatorname{sim}\left(\boldsymbol{z}_{t}^{n}, \boldsymbol{z}_{t}^{m}\right) / \tau}}{\sum_{j=1}^{N} e^{\operatorname{sim}\left(\boldsymbol{z}_{t}^{n}, \boldsymbol{z}_{t}^{j}\right) / \tau}},
\end{equation}
where $\operatorname{sim}(\boldsymbol{z}_{t}^{n}, \boldsymbol{z}_{t}^{m})$ is the cosine similarity $\frac{\boldsymbol{z}_{t}^{n\top} \boldsymbol{z}_{t}^{m}}{\left\|\boldsymbol{z}_{t}^{n}\right\| *\left\|\boldsymbol{z}_{t}^{m}\right\|}$, $\tau$ is a temperature hyper-parameter and $n$ is the number of sequences within a batch.

Although the training objective tries to pull representations with similar semantics closer and push dissimilar ones away from each other, these representations may still not be sufficiently discriminative and not be very robust to temporal noise. To demonstrate this, let us first denote angular $\theta_{n, m}$ as follows:
\begin{equation}
\theta_{n, m}=\arccos \left(\frac{\boldsymbol{z}_{t}^{n\top} \boldsymbol{z}_{t}^{m}}{\left\|\boldsymbol{z}_{t}^{n}\right\| * \|\boldsymbol{z}_{t}^{m}\| }\right)
\end{equation}
The angular margin for $\boldsymbol{z}_{t}^{n}$ in NT-Xent is $\theta_{n,m}=\theta_{n, j}$, as show in Fig.~\ref{Angular}. Due to the lack of a decision margin, a tiny time perturbation around the angular margin may lead to an incorrect decision. To overcome this problem, we propose a new training objective for temporal representation learning by adding an additive angular margin $\sigma$ between positive pairs $\boldsymbol{z}_{t}^{n}$ and $ \boldsymbol{z}_{t}^{m}$. This means that we want to keep some interval between the positive samples and do not force them exactly the same. We named it Time Angular Margin contrastive loss (TAM Loss), which can be formulated as follows:
\begin{equation}
    \mathcal{L}_{\text{TAM}}=-\log \frac{e^{\cos \left(\theta_{n,m}+\sigma \right) / \tau}}{e^{\cos \left(\theta_{n,m}+\sigma \right) / \tau}+\sum_{j \neq n} e^{\cos \left(\theta_{n,j}\right) / \tau}}
\end{equation}
The TAM loss introduces a angular margin for $\boldsymbol{z}_{t}^{n}$ that is defined as $\theta_{n,m}+\sigma =\theta_{n, j}$, as shown in Fig.~\ref{Angular}. Compared to the NT-Xent loss, the TAM loss further encourages $\boldsymbol{z}_{t}^{n}$ to move towards the region where $\theta_{n,m}$ is smaller and $\theta_{n, j}$ is larger. It increases the similarity of temporal representations with similar semantics and enlarges the discrepancy between different semantic representations. This enhances the alignment and uniformity properties, which are the two key measurements of representation quality related to contrastive learning~\cite{SimCSE}. Moreover, the angular margin provides an extra margin $\sigma$ to $\theta_{n,m}=\theta_{n, j}$, which is often utilized during inference, making the loss more tolerant to temporal noise and better at capturing the underlying semantic intentions of the user. Overall, these properties make the TAM loss a more effective training objective than traditional alternatives for extracting users' temporal intentions.

\subsubsection{Social Aware Block}
To capture the intrinsic spatial-temporal movement patterns of different users more effectively, we propose the Social Aware Block, which generates a unique representation for each user. Specifically, we utilize Graph Attention Networks (GATs) to aggregate the neighbor features of each user and employ an adaptive adjacency matrix to aggregate higher-order neighbor features.

Given the social network $\mathcal{G}=\{\mathcal{U, E}\}$, where $\mathcal{U}$ and $\mathcal{E}$ denote the user and link sets respectively, the $i$-th user is denoted as $u_i$. We denote the matrix $E_u\in \mathbb{R}^{\vert \mathcal{U} \vert\times D_u}$ as the lookup table of user embedding. Then we leverage GAT to aggregate neighbors' representations and update its embedding for each user, the new embedding is computed as:
\begin{equation}
    \boldsymbol{h}_{i}^{(l)}= \sigma\left(\sum\nolimits_{j \in \mathcal{N}_{i}} \alpha_{i j} \mathbf{W} \boldsymbol{h}_{j}^{(l-1)}\right),
\end{equation}
where $\boldsymbol{h}_{i}^{(l-1)}\in \mathbb{R}^{D_u}$ denotes the embedding of $u_i$ in the $(l-1)$-th layer. $\mathcal{N}_{i}$ is the neighbour set of user $i$. The initial embedding of user $u_i$, denoted by $\boldsymbol{h}_i^{(0)}$, is simply the $i$-th row of $E_u$. $\sigma$ is the sigmoid activation function. To compute the attention weight $\boldsymbol{a}_{ij}$ between users $u_i$ and $u_j$, we use the following formula:
\begin{equation}
    \alpha_{i j}=\frac{\exp \left(\varphi \left({\boldsymbol{a}}^{T}\left[\boldsymbol{h}_{i} \| \boldsymbol{h}_{j}\right]\right)\right)}{\sum_{c \in \mathcal{N}_{i}} \exp \left(\varphi \left({\boldsymbol{a}}^{T}\left[\boldsymbol{h}_{i} \| \boldsymbol{h}_{c}\right]\right)\right)}
\end{equation}
where $\varphi$ is the LeakyReLu activation function. Finally, we concat the final user embedding $\boldsymbol{h}_{u}$ with $\boldsymbol{z}_t$ to update the user's temporal representation $\boldsymbol{z}_t$.

\subsection{ST Cross-View Contrastive Module}
The temporal intention and the spatial topic are strongly correlated with users and highly susceptible to each other's impacts. This gives rise to a wealth of high-quality self-supervised signals. For example, most users engage in activities such as eating three meals, working, and exercising, but their schedules vary between weekdays and weekends and have different spatial topics. Different professions exhibit diverse spatial topics during the week due to their unique requirements. For instance, doctors, service workers, and students have distinct temporal intentions owing to their professional requirements. 

\begin{figure}[htbp]
    \centering
    \includegraphics[width=1\columnwidth]{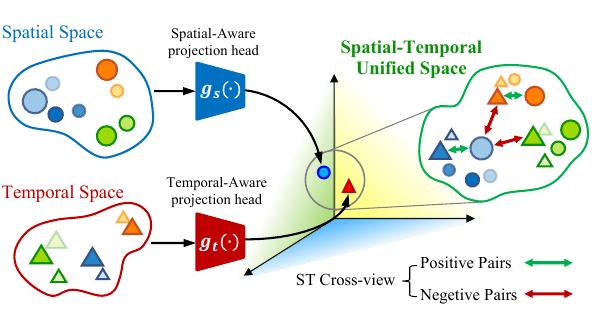}
    \caption{The architecture of Spatial-Temporal Cross-view Module. The circle represents the spatial representation and the triangle represents the temporal representation. The similar color represents the representations from the same check-in sequence that the deep one is the anchor and the other is its augmentation.}
    \label{TKDE_Gong_Cross}
\end{figure}

To model check-in sequences based on these self-supervised cross-view signals, we propose a spatial-temporal cross-view contrastive learning framework. This approach enables the encoder to focus on learning optimal representations in both temporal and spatial views in the early stage and fuse at the semantic level. As learning progresses, we align the temporal and spatial representations by unifying them into a shared semantic space based on spatial-temporal parallel pairs. By tightly fusing the most relevant semantic information from the temporal and spatial views, we fully utilize the diverse check-in sequence data to uncover the spatial-temporal patterns of users.

The Spatial-Temporal Cross-View contrastive module is designed to learn unified check-in sequence representations prior to fusion. It learns a similarity function $s=g_{s}\left(\boldsymbol{z}_s\right)^{\top} g_{t}\left(\boldsymbol{ z}_t\right)$, which assigns higher similarity scores to parallel spatial-temporal pairs. Here, $g_{s}$ and $g_{t}$ are spatial and temporal projection heads that map the representation of sequences to a unified semantic space. For each spatial and temporal pair, we calculate the softmax-normalized spatial-to-temporal similarity score as follows:
\begin{equation}
    \boldsymbol{p}_{m}^{\mathrm{s} 2 t}(\boldsymbol{z}_s)=\frac{\exp \left(s\left(\boldsymbol{z}_s, \boldsymbol{ z}_t^m\right) / \tau\right)}{\sum_{n} \exp \left(s\left(\boldsymbol{z}_s,\boldsymbol{ z}_t^n\right) / \tau\right)}, 
\end{equation}
and the temporal-to-spatial similarity as:
\begin{equation}
    \boldsymbol{p}_{m}^{t 2 \mathrm{s}}(\boldsymbol{ z}_t)=\frac{\exp \left(s\left(\boldsymbol{ z}_t, \boldsymbol{z}_s^m
    \right) / \tau\right)}{\sum_{n} \exp \left(s\left(\boldsymbol{ z}_t, \boldsymbol{z}_s^n\right) / \tau\right)}, 
\end{equation}
where $\tau$ is a learnable temperature parameter. Let $\boldsymbol{y}^{\mathrm{s} 2 t}(\boldsymbol{z}_s)$ and $\boldsymbol{y}^{t 2 \mathrm{s}}(\boldsymbol{ z}_t)$ denote the ground-truth one-hot similarity, where negative pairs have a probability of 0 and the positive pair has a probability of 1. The spatial-temporal contrastive loss is defined as the cross-entropy H between $\boldsymbol{p}$ and $\boldsymbol{y}$ :
\begin{equation}  
\resizebox{.8\linewidth}{!}{
        \begin{math}
        \begin{aligned}
            \mathcal{L}_{ST}=\frac{1}{2} \mathbb{E}_{(\boldsymbol{z}_s, \boldsymbol{ z}_t) \sim D}[\mathrm{H}&\left(\boldsymbol{y}^{s2t}(\boldsymbol{z}_s), \boldsymbol{p}^{s2t}(\boldsymbol{z}_s)\right) \\&+\mathrm{H}\left(\boldsymbol{y}^{t2s}(\boldsymbol{ z}_t), \boldsymbol{p}^{t2s}(\boldsymbol{ z}_t)\right)]
        \end{aligned}
        \end{math}
    }
\hspace{2em}
\end{equation}
Ultimately, the pre-training loss is represented as:
\begin{equation}
    \mathcal{L}_{Pre} =\mathcal{L}_{Spatial}+\mathcal{L}_{TAM}+\mathcal{L}_{ST}.
\end{equation}

\subsection{Fine-tuning for Downstream Applications}
We employ the training set to pre-train the STCCR. We combine the spatial representation with the temporal representation as the global human behavior representation during the pre-training stage. Then we use a projection head to fine-tune among next location prediction (\textbf{LP}), next time prediction (\textbf{TP}), and trajectory user link (\textbf{TUL}) tasks, respectively. Firstly, we consider LP and TUL downstream tasks as multiclassification problems, as formulated below. Given a check-in sequence $\mathcal{T}^u$ from a specific user $u\in|\mathcal{U}|$, we feed it to the pre-trained encoder to obtain the check-in sequence representation $G(\mathcal{T}^{u})$. Then we use a projection head $f_{\boldsymbol\theta}$ to predict the classification $y$ such as the next location where the user will soon arrive or the user who generated this check-in sequence, \emph{i.e.}, $f_{\boldsymbol\theta}\left(G(\mathcal{T}^{u}), \boldsymbol\theta\right) \mapsto y$. We maximize the conditional log-likelihood $\log f_{\boldsymbol\theta}( y\mid G(\mathcal{T}^{u}))$ for a given $N$ observations $\left\{\left(G(\mathcal{T}^{u}),y\right)\right\}_{i=1}^{N}$ as follows:
\begin{equation}
\label{eq_downstream}
\resizebox{.80\linewidth}{!}{
    \begin{math}
    \begin{aligned}
        \mathcal{L}_{M L E}(\boldsymbol\theta) &=\sum_{i=1}^{N} \log{f_{\boldsymbol\theta}\left( y \mid G(\mathcal{T}^{u})\right)}, \\
        f_{\boldsymbol\theta}\left( y \mid G(\mathcal{T}^{u})\right) &= \;\mathrm{softmax} \left(\boldsymbol W G(\mathcal{T}^{u})+\boldsymbol{b}\right).
    \end{aligned}
    \end{math}
}
\end{equation}
%

Secondly, for downstream tasks of time prediction, we follow the method of IFLTPP~\cite{IFLTPP} using an intensity-free method to model the interaction time as a mixture distribution. We first gain the mixture weights $ w$, means $\mu$ and standard deviations ${s}$ from the check-in sequence representation $G(\mathcal{T}^{u})$. Then we build a mixed distribution function and sample to get the prediction time $\tau$ as follows:

\begin{equation}
\label{eq_tp_distribution}
p\left( \tau \mid w, \mu, s \right) = \sum_{k=1}^{K}\frac{1}{\tau s_k \sqrt{2\pi}}\exp{-\frac{(\log{\tau} - \mu_k)^2}{2{s_k}^2}},
\end{equation}
where $k$ represents the number of independent Gaussian distributions in the mixed distribution.
Then, we can sample from the mixture model in the parsing solution.
\begin{equation}
\label{eq_tp_tau}
\tau = \sum_{k=1}^K{{w_k}\exp{(a{\mu_k} + b + \frac{a^2{s_k}}{2})}},
\end{equation}
where $a$ denotes the mean of the whole set and $b$ denotes the standard deviation of the whole set.
\section{Experiments}
To evaluate the performance of our proposed model, we carried out extensive experiments on three real-world check-in sequence datasets, targeting three different types of downstream: Location Prediction (LP), Trajectory User Link (TUL), and Time Prediction (TP). \textit{The code has been released at: https://github.com/LetianGong/STCCR}.
\subsection{Datasets}
\label{sec:datasets}
In our experiments, we use three real-world datasets derived from raw \newrevise{WeePlace~\footnote{\newrevise{http://www.yongliu.org/datasets}}, Gowalla~\footnote{\newrevise{https://snap.stanford.edu/data/loc-Gowalla.html}} of New York City (NYC), and Tokyo (TKY) check-in data}. Our model undergoes a filtering process that selects high-quality trajectory sequences for training. To ensure data consistency, we set a maximum historical time limit of 120 days and filter out users with fewer than 10 records and places visited fewer than 10 times. Table~\ref{datasets} provides statistical information for each proceed dataset.
We split the datasets into training, validation, and test sets at a 6:2:2 ratio. The three datasets exhibit distinct spatial-temporal correlations. For example, on workdays, a user may visit a breakfast restaurant at a certain time, while on rest days, the user's eating schedule may shift, making their behavior more time-sensitive. Due to the large number of POIs, sparse data sets, irregular time intervals, and varying user intentions, it is challenging to forecast and extract geographic and temporal information. Through our experiments, we demonstrate the superiority of our STCCR model, which we evaluate on all three datasets.
\begin{table}[h]
        \centering
        \caption{Statistics of datasets}
        \label{datasets}
        \begin{tabular}{ccccc}
        \hline
            & \text { Gowalla-NYC } & \text { Gowalla-TKY } & \text { WeePlace }  \\
            \hline \text { \#users } & 2,333 & 3,583 & 1,028  \\
            \text { \#locations } & 7,690 & 46,278 & 9,295 \\
            \text { \#check-ins } & 74,022 & 128,916 & 104,762 \\
            \hline
        \end{tabular}
     
\end{table}

\subsection{Baselines}
\paratitle{Location Prediction Methods}
We cover one classic check-in prediction models and three state-of-the-art LP models to demonstrate the superiority of our model.
    \begin{itemize}
    
        \item \textbf{DeepMove} ~\cite{DeepMove} is a classical check-in sequence position prediction model to capture the periodicity of trajectory motion.
         \item \textbf{PLSPL}~\cite{PLSPL} learns the long-term preference of the user by attention and the short-term preference of two LSTMs.
        \item \textbf{LightMove}~\cite{LightMove} leverages neural ordinary differential equations to enhance resilience against sparse or incorrect input.
         \item \textbf{HMT-LSTM}~\cite{HMT-GRN} addresses data sparsity by learning user region matrices with varying lower sparsities.
    \end{itemize}
    
\paratitle{Trajectory User Link Methods}
We select two end-to-end and two pre-trained TUL task models for comparison.
     \begin{itemize}
        \item \textbf{TULER}~\cite{TULER} is the first study to propose the trajectory user link and simulate it using RNN.
        \item \textbf{TULVAE} ~\cite{TULVAE} follows the work of TULER and pre-trains the encoder by adding VAE on a prior basis.
        \item \textbf{MoveSim}~\cite{MoveSim} captures the temporal changes in human motion employing a generative adversarial network for trajectory pre-training.
        \item \textbf{S2TUL} ~\cite{S2TUL} combines homogeneous graphs and sequential neural networks to a greedy method to relink trajectories.
    \end{itemize}
\paratitle{Time Prediction Methods}
We selected one classic and two SOTA models for comparison.
\begin{itemize}
    \item \textbf{IFLTPP}~\cite{IFLTPP} shows how to overcome the limitations of intensity-based approaches by directly modeling the conditional distribution of inter-event times. It draws on the literature on normalizing flows to design models that are flexible and efficient
    \item \textbf{THP}~\cite{THP} extends the transformers to include time and mark influences between events to calculate.
the conditional intensity function for the arrival of future events in the sequence
     \item \textbf{NSTPP}~\cite{NSTPP} acquires the ability to retrieve and prioritize a relevant set of continuous-time event sequences based on a given anchor sequence.
\end{itemize}
\paratitle{Sequence Representation Methods}
We select four representative baselines for the representation of contrast sequences using contrastive learning. We apply them to learn the representation of the check-in sequence and serve three downstream tasks.
\begin{itemize}
    \item \textbf{ReMVC}~\cite{ReMVC} designs an intra-inter view contrastive learning module to learn distinct region embeddings and transfer knowledge across multiple views.
    \item \textbf{VaSCL}~\cite{VaSCL} is suggested to add hard negative samples into the NLP field. It uses smaller batches to get better performance.
     \item \textbf{SML}~\cite{SML} used self-supervised representation to harmonize noisy and nonuniform length trajectories.
     \item \textbf{CACSR}~\cite{CACSR} utilizes adversarial perturbations in contrastive learning for data augmentation to enhance the capability of sequence representation.
\end{itemize}

\begin{table*}[!t]
	\centering 
     \caption{Next location prediction (LP) and trajectory user link (TUL) performance comparison between different approaches.}
	\scalebox{0.92}{
	\begin{threeparttable}
		\begin{tabular}{c|c|ccc|ccc|ccc}
			\toprule
			\multicolumn{2}{c|}{Datasets} &
              \multicolumn{3}{c|}{\textbf{Gowalla-NYC}} &
              \multicolumn{3}{c|}{\textbf{Gowalla-TKY}} &
              \multicolumn{3}{c}{\textbf{WeePlace}} \\ \midrule
            \multicolumn{2}{c|}{Metric} &
              \multirow{2}{*}{Acc@5 (\%)$\uparrow$} &
              \multirow{2}{*}{Acc@20 (\%)$\uparrow$} & \multirow{2}{*}{MRR (\%)$\uparrow$} &
              \multirow{2}{*}{Acc@5 (\%)$\uparrow$} &
              \multirow{2}{*}{Acc@20 (\%)$\uparrow$} & \multirow{2}{*}{MRR (\%)$\uparrow$} &
              \multirow{2}{*}{Acc@5 (\%)$\uparrow$} &
              \multirow{2}{*}{Acc@20 (\%)$\uparrow$} & \multirow{2}{*}{MRR (\%)$\uparrow$} \\ \cline{1-2}
              \multirow{1}{*}[-2pt]{Task} &
              \multirow{1}{*}[-2pt]{Method} & & & & & & & & & \\
			\midrule
            & DeepMove & 34.04$\pm$0.16 & 51.70$\pm$0.29 & 28.88$\pm$0.41 & 20.73$\pm$0.35 & 29.01$\pm$0.12 & 15.24$\pm$0.37 & 34.39$\pm$0.18 & 47.37$\pm$0.28 & 25.25$\pm$0.23 \\ 
            ~ & LightMove & 36.75$\pm$0.31 & 53.78$\pm$0.18 & 30.72$\pm$0.17 & 23.11$\pm$0.30 & 33.45$\pm$0.23 & 17.73$\pm$0.34 & 37.91$\pm$0.23 & 54.49$\pm$0.21 & 28.76$\pm$0.22 \\ 
            ~ & PLSPL & 36.93$\pm$0.31 & \underline{54.42$\pm$0.38} & 30.98$\pm$0.43 & \underline{23.17$\pm$0.37} & 33.47$\pm$0.39 & 17.49$\pm$0.21 & 38.82$\pm$0.44 & 55.19$\pm$0.23 & 29.47$\pm$0.21 \\ 
            ~ & HMT-LSTM & 36.57$\pm$0.21 & 53.62$\pm$0.38 & 30.41$\pm$0.43 & 22.48$\pm$0.33 & 32.99$\pm$0.19 & 17.07$\pm$0.42 & 37.84$\pm$0.19 & 54.14$\pm$0.25 & 28.41$\pm$0.19 \\ 
              \textbf{LP}
            ~ & VaSCL & 29.04$\pm$0.18 & 40.70$\pm$0.18 & 22.88$\pm$0.46 & 18.73$\pm$0.23 & 27.01$\pm$0.14 & 14.24$\pm$0.23 & 30.39$\pm$0.29 & 43.37$\pm$0.30 & 23.25$\pm$0.41 \\ 
            ~ & ReMVC & \underline{37.61$\pm$0.19} & 54.35$\pm$0.22 & \underline{31.07$\pm$0.41}  & 23.15$\pm$0.22 & \underline{33.61$\pm$0.29} & \textbf{18.09$\pm$0.34} & \underline{39.03$\pm$0.43} & \underline{55.21$\pm$0.16} & 28.94$\pm$0.29 \\ 
            ~ & SML & 35.75$\pm$0.15 & 53.88$\pm$0.22 & 30.15$\pm$0.15 & 21.56$\pm$0.35 & 31.63$\pm$0.32 & 16.82$\pm$0.31 & 37.87$\pm$0.26 & 53.19$\pm$0.28 & 27.75$\pm$0.23 \\ 
            ~ & CACSR & 33.35$\pm$0.44 & 51.05$\pm$0.15 & 29.24$\pm$0.17 & 23.12$\pm$0.37 & 33.57$\pm$0.39 & 17.89$\pm$0.21 & 39.02$\pm$0.41 & 55.20$\pm$0.16 & \underline{29.64$\pm$0.29} \\ 
            ~ & \textbf{STCCR} & \textbf{38.38$\pm$0.15} & \textbf{54.78$\pm$0.24} & \textbf{31.24$\pm$0.25} & \textbf{23.62$\pm$0.23} & \textbf{34.43$\pm$0.34} & \underline{17.93$\pm$0.23} & \textbf{39.16$\pm$0.16} & \textbf{55.36$\pm$0.32} & \textbf{30.07$\pm$0.11} \\ 
         
            \midrule
            & TULER & 59.71$\pm$0.23 & 69.15$\pm$0.20 & 54.17$\pm$0.18 & 70.87$\pm$0.28 & 79.25$\pm$0.26 & 66.74$\pm$0.42 & 73.61$\pm$0.29 & 80.98$\pm$0.46 & 70.88$\pm$0.35 \\ 
            ~ & TULVAE & 60.94$\pm$0.19 & 69.56$\pm$0.19 & 56.01$\pm$0.18 & 73.19$\pm$0.31 & 79.46$\pm$0.31 & 67.52$\pm$0.29 & 75.78$\pm$0.27 & 86.43$\pm$0.18 & 72.92$\pm$0.41 \\ 
            ~ & MoveSim & 64.21$\pm$0.32 & 72.12$\pm$0.18 & 59.56$\pm$0.12 & 72.12$\pm$0.41 & 79.78$\pm$0.16 & 67.23$\pm$0.22 & 82.35$\pm$0.18 & 87.32$\pm$0.15 & 74.27$\pm$0.21 \\ 
            ~ & S2TUL & \underline{65.49$\pm$0.25} & 73.21$\pm$0.32 & 60.19$\pm$0.19 & \underline{75.04$\pm$0.21} & \underline{81.68$\pm$0.35} & \underline{70.64$\pm$0.34} & 83.46$\pm$0.19 & \underline{89.33$\pm$0.23} & \underline{77.82$\pm$0.29} \\ 
            \textbf{TUL}
            & VaSCL & 55.43$\pm$0.15 & 65.54$\pm$0.19 & 48.43$\pm$0.37 & 53.94$\pm$0.36 & 57.23$\pm$0.29 & 53.15$\pm$0.43 & 62.15$\pm$0.29 & 70.83$\pm$0.21 & 54.45$\pm$0.17 \\ 
            ~ & ReMVC & 65.37$\pm$0.37 & \underline{73.23$\pm$0.22} & \underline{60.34$\pm$0.33} & 74.23$\pm$0.46 & 81.67$\pm$0.28 & 69.34$\pm$0.45 & \underline{83.52$\pm$0.16} & 88.72$\pm$0.18 & 76.92$\pm$0.36 \\ 
            ~ & SML & 64.57$\pm$0.21 & 72.62$\pm$0.38 & 59.41$\pm$0.43 & 72.48$\pm$0.33 & 79.99$\pm$0.19 & 67.07$\pm$0.42 & 82.84$\pm$0.19 & 87.14$\pm$0.25 & 74.41$\pm$0.19 \\ 
            ~ & CACSR & 63.75$\pm$0.15 & 72.88$\pm$0.22 & 59.15$\pm$0.15 & 73.15$\pm$0.22 & 80.61$\pm$0.29 & 68.29$\pm$0.34 & 82.03$\pm$0.43 & 88.29$\pm$0.16 & 76.94$\pm$0.29 \\ 
            ~ & \textbf{STCCR} & \textbf{66.17$\pm$0.11} & \textbf{74.04$\pm$0.13} & \textbf{61.79$\pm$0.08} & \textbf{75.48$\pm$0.31} & \textbf{82.08$\pm$0.25} & \textbf{70.97$\pm$0.29} & \textbf{84.58$\pm$0.08} & \textbf{90.14$\pm$0.05} & \textbf{78.03$\pm$0.36} \\ 
            \bottomrule
        \end{tabular}
        

    \end{threeparttable}
    }

\label{table:baseline-results_1}
\end{table*}

\begin{table*}[!t]
	\centering 
     \caption{Next time prediction (TP) performance comparison between different approaches.}
	\scalebox{0.95}{
	\begin{threeparttable}
		\begin{tabular}{c|c|ccc|ccc|ccc}
			\toprule
			\multicolumn{2}{c|}{Datasets} &
              \multicolumn{3}{c|}{\textbf{Gowalla-NYC}} &
              \multicolumn{3}{c|}{\textbf{Gowalla-TKY}} &
              \multicolumn{3}{c}{\textbf{WeePlace}} \\ \midrule
            \multicolumn{2}{c|}{Metric} &
              \multirow{2}{*}{MAE$\downarrow$} &
              \multirow{2}{*}{RMSE$\downarrow$} & \multirow{2}{*}{NLL(e-2)$\downarrow$} &
              \multirow{2}{*}{MAE$\downarrow$} &
              \multirow{2}{*}{RMSE$\downarrow$} & \multirow{2}{*}{NLL(e-2)$\downarrow$} &
              \multirow{2}{*}{MAE$\downarrow$} &
              \multirow{2}{*}{RMSE$\downarrow$} & \multirow{2}{*}{NLL(e-2)$\downarrow$} \\ \cline{1-2}
              \multirow{1}{*}[-2pt]{Task} &
              \multirow{1}{*}[-2pt]{Method} & & & & & & & & & \\
			\midrule

            & IFLTPP & 25.34 $\pm$0.19 & 36.93 $\pm$0.26 & 66.91 $\pm$0.27 & 22.98 $\pm$0.36 & 33.21 $\pm$0.19 & 60.13 $\pm$0.18 & 29.21 $\pm$0.28 & 36.92 $\pm$0.18 & \textbf{74.92 $\pm$0.32} \\ 
            ~ & THP & 24.77 $\pm$0.23 & 35.07 $\pm$0.20 & 66.94 $\pm$0.16 & 21.97 $\pm$0.31 & 32.91 $\pm$0.3 & \textbf{59.09 $\pm$0.32} & \underline{27.66 $\pm$0.36} & \underline{35.30 $\pm$0.27} & \underline{74.98 $\pm$0.21} \\ 
            ~ & NSTPP & \underline{24.32 $\pm$0.18} & \underline{35.02 $\pm$0.21} & \underline{66.84 $\pm$0.26} & \underline{21.37 $\pm$0.26} & \underline{32.48 $\pm$0.17} & 61.87 $\pm$0.25 & 28.24 $\pm$0.29 & 36.09 $\pm$0.18 & 78.36 $\pm$0.25 \\ 
            ~ & VaCSL & 29.09 $\pm$0.16 & 37.99 $\pm$0.31 & 79.38 $\pm$0.15 & 23.70 $\pm$0.21 & 35.92 $\pm$0.27 & 65.23 $\pm$0.33 & 31.86 $\pm$0.22 & 38.48 $\pm$0.27 & 85.63 $\pm$0.15 \\
            \textbf{TP}
            & ReMVC & 25.86 $\pm$0.21 & 37.14 $\pm$0.23 & 71.11 $\pm$0.22 & 22.48 $\pm$0.23 & 33.60 $\pm$0.19 & 62.55 $\pm$0.33 & 28.51 $\pm$0.33 & 36.17 $\pm$0.27 & 80.07 $\pm$0.15 \\ 
            ~ & SML & 26.28 $\pm$0.17 & 36.03 $\pm$0.25 & 71.52 $\pm$0.26 & 22.78 $\pm$0.31 & 33.42 $\pm$0.22 & 62.26 $\pm$0.22 & 29.47 $\pm$0.31 & 35.88 $\pm$0.18 & 77.75 $\pm$0.15 \\ 
            ~ & CACSR & 27.05 $\pm$0.24 & 37.69 $\pm$0.21 & 73.73 $\pm$0.35 & 22.67 $\pm$0.32 & 34.07 $\pm$0.28 & 64.13 $\pm$0.37 & 30.71 $\pm$0.15 & 38.34 $\pm$0.24 & 82.84 $\pm$0.32 \\ 
            ~ & \textbf{STCCR} & \textbf{24.25 $\pm$0.22} & \textbf{34.68 $\pm$0.27} & \textbf{64.86 $\pm$0.28} & \textbf{20.99 $\pm$0.25} & \textbf{31.87 $\pm$0.35} & \underline{59.16 $\pm$0.24} & \textbf{27.28 $\pm$0.31} & \textbf{35.28 $\pm$0.31} & 75.41 $\pm$0.24 \\ 
            \bottomrule
        \end{tabular}
        

    \end{threeparttable}%

}
\label{table:baseline-results_2}
\end{table*}
\subsection{Settings}
{We standardize all data using the Log Mean-Std approach and feed the normalized data into the model, which is optimized using reverse mode automatic differentiation and Adam \cite{kingma2014adam}. To train the model, historical data of locations and users are first embedded, followed by location features fed into the spatial topic module and user IDs fed into the user embedding layer. We found that the time information worked better without the embedding layer, so we did not embed it. The representation vectors output from the spatial topic module and the temporal intention module are first mapped in the spatial-temporal joint space by the projection head. After that, we do contrastive learning between the outcomes of the two modules. Three separate downstream tasks are subsequently completed by combining the two representations from those two modules. 

To assess the projected values, we retransform them back to the actual values and compare them to the ground truth. The evaluation measures of the location prediction and trajectory user link tasks include Acc@k and mean reciprocal rank (MRR). The evaluation measures of the time prediction task include mean absolute error (MAE), the root mean square error (RMSE), and mean absolute percentage error (MAPE). The STCCR model was constructed using the PyTorch\footnote{https://pytorch.org}. The loss function is a cross-entropy loss for the LP and TUL tasks and an MAE loss for the TP task. The performance on the validation sets determines the hyper-parameters and the best models. All experiments are performed five times, and the means and standard deviations are calculated. To make a fair comparison, for all methods, the embedding dimension $d$ is 256, while the hidden state $h$ has a size of 256. The learning rate is 0.001. Other detailed settings of the STCCR model for the three datasets are described in~\ref{datasets}. The is pre-trained for 100 epochs on the training sets with the early-stopping mechanism of 10 patience. All trials have been conducted on Intel Xeon E5-2620 CPUs and NVIDIA RTX A5000 GPUs.
}

\subsection{Comparison and Analysis of Results on the DownStream Tasks}
Table~\ref{table:baseline-results_1} and Table~\ref{table:baseline-results_2} shows the comparison results of our model with other baseline models on three downstream tasks. The best is shown in \textbf{bold}, and the second-best is shown as \underline{underlined}. \par
Our representation learning pre-trained models can meet or even exceed the best-performing end-to-end models. Besides, the performance of STCCR far outperforms other sequential representation learning models on three tasks. In the LP task, STCCR improves the sota representation models by 3.9\%, 2.5\% 7.1\%  in terms of Acc@5, Acc@20, and MRR average on the three datasets, respectively.  In the TUL task, STCCR improves by 4.3\%, 4.7\%, 5.3\%  in terms of Acc@5, Acc@20, and MRR average on the three datasets, respectively. In the TP task, STCCR improves by  7.9\%, 4.4\%, 1.3\% in terms of MAE, RMSE, and NLL average on the three datasets, respectively.
These results demonstrate that our model performs very well on these downstream tasks. The strength of our model benefits from our spatial-temporal cross-view contrastive framework, which is able to encourage the spatial topic module and temporal intention module to mine the spatial-temporal patterns from mobility behavior data. The spatial topic module uses contrastive clustering can effectively capture the spatial semantics of human mobility and enhance the generalization of the model by clustering shared mobility patterns.  The temporal intention module uses angular contrastive manner which can force the same sample to be not exactly the same, but within a small region. This is a good way to mitigate the effects of temporal uncertainty and noise on the user's temporal intention. The self-supervised signals from unified space can effectively mitigate the sparsity of check-in data and spur the encoder to learn accurate general representations for check-in sequences. In general, the superiority and versatility of our method show that the spatial-temporal cross-view contrastive framework proposed in this work is well suited for modeling check-in sequences.

ReMVC designs an intra-view and an inter-view contrastive learning module to transfer knowledge across multiple views, but it does not effectively extract the semantic information before the information interaction.  SML used a prior fusion manner to learn representation for every check-in sequence but ignoring the diversity of user movement and the temporal noise can't capture the potential patterns of activity. The performance of VaSCL on the check-in dataset conclusively proves that data augmentation techniques in NLP are not directly transferable to human movement trajectory data. Although difficult negative samples are introduced in VaSCL. The contrastive method of similar semantic samples is considered in the contrastive process, but it is still unable to mine the spatial topic and temporal intention of users based on the characteristics of check-in data. Our previous work, CACSR, performs data augmentation by combating perturbations, which can largely combat the noise in the original data, but ignore the shared topic of users moving in spatial.

\subsection{Ablation experiments}
\begin{figure}[h]
    \centering
    \includegraphics[width=1\columnwidth]{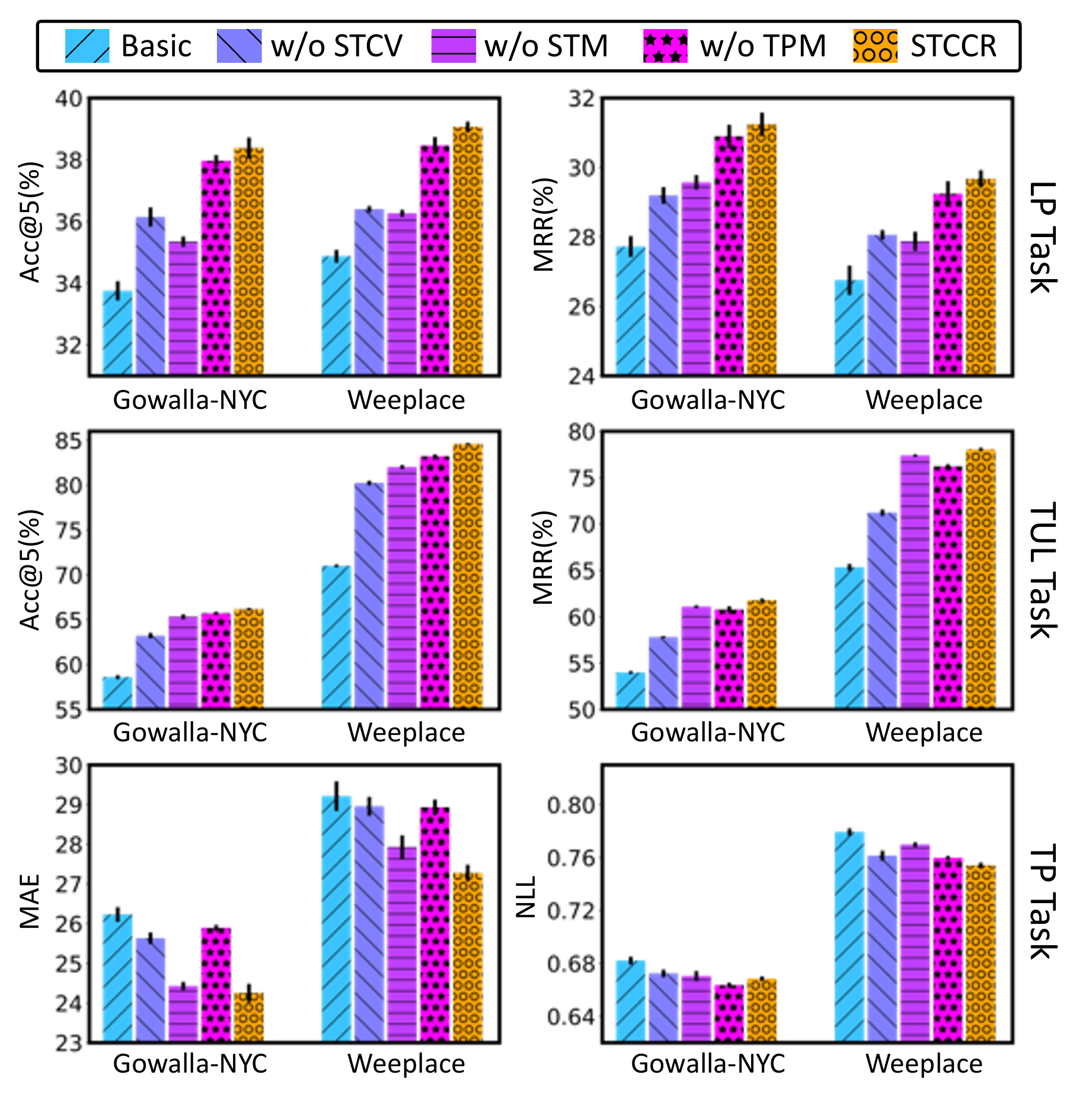}
    \caption{Component analysis of STCCR}
    \label{Fig_ablation}
\end{figure}

To further evaluate the effects of different components in STCCR, we conduct ablation experiments and analyze experimental results on the Gowalla-NYC and WeePlace datasets. We compare these four variants with the STCCR model for the three downstream tasks. Fig.~\ref{Fig_ablation} shows the results.

     \begin{itemize}
        \item \textbf{Basic:} We use contrastive manners like SimCSE and InfoNCE loss to replace the origin manner in Spatial Topic Module and Temporal Intention Module. Simply combine two representation vectors to complete the downstream task.
        \item \textbf{w/o STCV:} We remove the ST Cross-View Module, instead we directly combine the representation from the other two modules. The rest of the settings are the same as STCCR. We use this setting to evaluate the function of the ST Cross-View Module.
        \item \textbf{w/o STM:} We use contrastive manner like SimCSE and InfoNCE loss to replace the origin manner in only Spatial Topic Module. The rest of the settings are the same as STCCR. The rest of the settings are the same as STCCR. We use this setting to evaluate the function of the Spatial Topic Module.
         \item \textbf{w/o TIM:} We use a contrastive manner like SimCSE and InfoNCE loss to replace the origin manner in only the Temporal Intention Module. The rest of the settings are the same as STCCR. The rest of the settings are the same as STCCR. We use this setting to evaluate the function of the Temporal Intention Module.
    \end{itemize} 
Among them, the spatial topic module has the greatest improvement for the next location prediction task. This is thanks to the spatial topic module, which combines the advantages of contrastive learning and clustering methods to effectively capture the shared spatial patterns among different users. This approach avoids learning a representation for each sample and can be valid for mitigating inadequate generalizability. This can motivate the model to more effectively mine higher-level semantic information about users' spatial movements from clustered self-supervised signals.

The temporal intentions module has the greatest improvement for the time prediction task. This is because the temporal intention module combines the advantages of contrastive learning and decision boundaries. By introducing decision bounds, the noise in the time series can be well filtered. This alleviates the difficulties in mining users' temporal intentions on check-in datasets with large temporal uncertainty.

The ST cross-view contrastive module has significant improvements for the trajectory user task and the next location prediction task. The ST cross-view contrastive module can exploit the natural self-supervised signals between temporal and spatial. It effectively mines the historical behavioral habits and neighboring travel intentions of users. It can be seen that the introduction of this module has a huge effect on the overall boost of the model.
    
This result proves the usefulness of three modules in using contrastive clustering, angular margin, and ST cross-view contrastive manner to learn the representation of check-in sequences. Finally, these modules are combined in our final model to produce the best results.

\begin{figure*}[htb]
	\centering
	\includegraphics[width=1\linewidth]{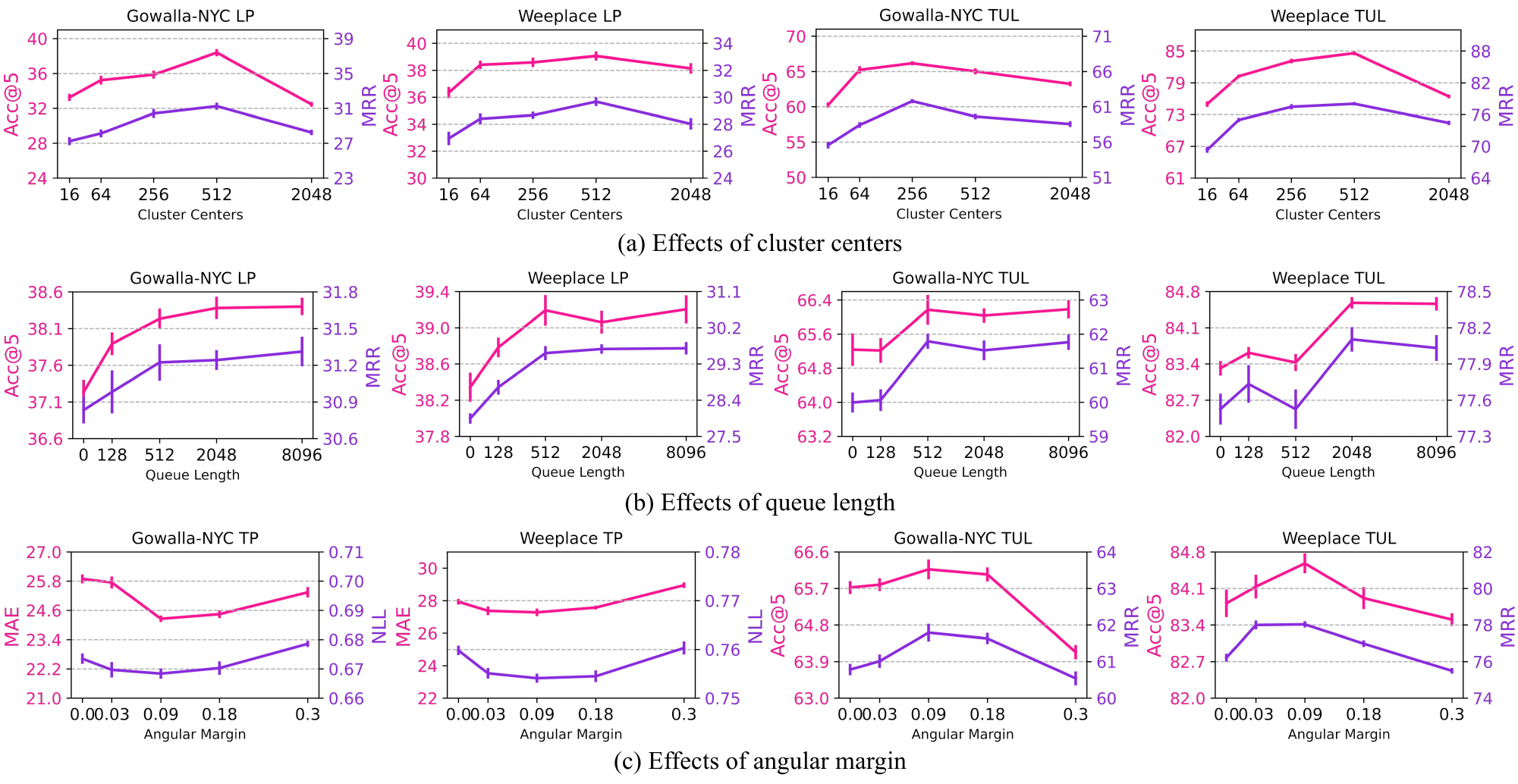}
	\caption{Effects of hyper-parameters validated on Gowalla-NYC and Weeplace dataset.}
	\label{hyper}
\end{figure*}

\subsection{Effects of Parameters}
In this section, we evaluate the effects of hyper-parameters in three modules: 1) the cluster number and queue length in the STM. 2) The theta margin during angular contrast in the TIM. 3) The loss weight of the ST cross-view module during the pre-training. The experiments are conducted on all the datasets, and while evaluating one of the hyper-parameters, we lock the other ones to optimum. We set the best number of cluster centers for each dataset in Table~\ref{tab:optimal-parameter}.

\subsubsection{Effects of Cluster Center Number}
Fig.~\ref{hyper}a shows the experimental results on the hyper-parameter tuning of cluster number. 
From these figures, we observe that the performance first improves as we increase the number of cluster centers, then deteriorates as it exceeds the optimum point. A small number of centers means thousands of users only have a limited number of spatial topics, which fails to capture the diversity of human activities. A large number of centers will cluster too many activity themes. This will make the method fail to extract shared activity themes. This degenerates into a w/o STM among the ablation experiments, i.e., learning a sequence representation for each sample. Reduces the generalization performance of the STM module. A moderate number of cluster centers can better capture semantic information about user movement in spatial. 

\subsubsection{Effects of Queue Length}
Fig.~\ref{hyper}b shows the experimental results on the hyperparameter tuning of the queue length. This figure demonstrates that performance improves steadily as we increase the length of the queue in the STM. A longer queue can contain more samples, nourishing the cluster centers in each epoch and helping downstream tasks gain better results. Yet, we observe that the degree of performance improvement is limited when the queue length is longer than eight thousand, and the longer queue will lead to a higher computational expense.

\begin{table}[t]
    \centering
    \caption{Settings of the STCCR model in three datasets.}
    \label{tab:optimal-parameter}
    \begin{threeparttable}
        \begin{tabular}{c|c}
            \toprule
            \begin{minipage}{0.2\columnwidth}\centering Parameter\end{minipage} & \begin{minipage}{0.32\columnwidth}\centering Range\end{minipage} \\
            \midrule
            Cluster centers & 16, 64, 256, \underline{512}, 2048 \\
            Queue length & 0, 128,512,\underline{2048},8096 \\
            Angular margin & 0, 0.03, \underline{0.09}, 0.18,0.3 \\
            Projection head size & 32, 128, \underline{512}, 2048 \\
            \bottomrule
        \end{tabular}
        \begin{tablenotes}
            \item{\underline{Underline} denotes the optimal value.}
        \end{tablenotes}
    \end{threeparttable}
\end{table}
\begin{figure}[htb]
	\centering
	\includegraphics[width=1\columnwidth]{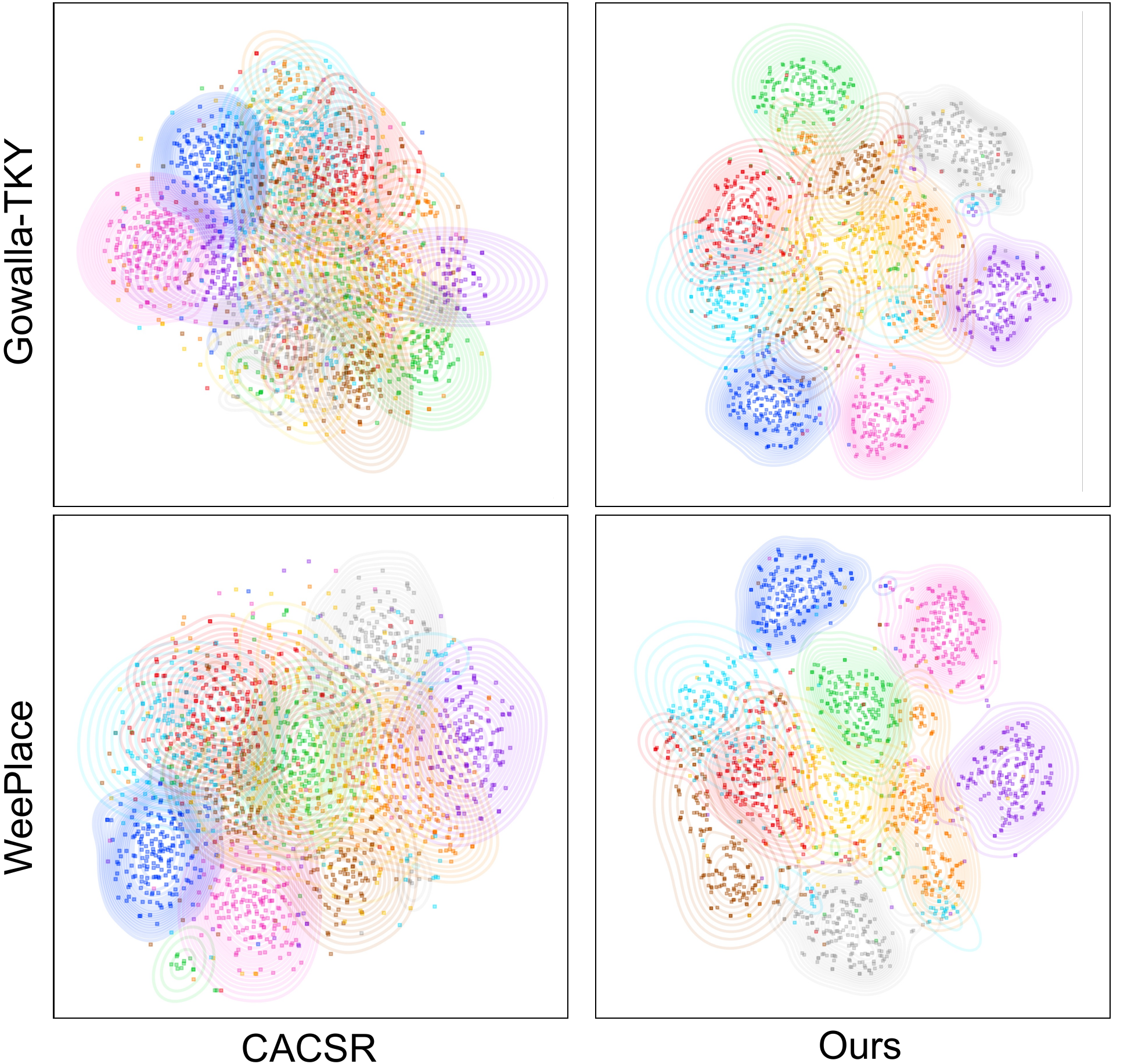}
	\caption{Case visualization comparison between CACSR and Ours.}
	\label{case_study}
\end{figure}
\subsubsection{Effects of  Theta Margin}
Fig.~\ref{hyper}c shows the experimental results on the hyperparameter tuning influence of the theta margin. It can be seen that theta margin has a great impact on the TP task. A too-small margin for noise does not play a filtering role; a large margin will lead to meaningless comparison training. Therefore, a suitable theta margin can effectively filter the noise while capturing the macroscopic temporal intention. In conclusion, we set the theta margin to 0.09 for all datasets.
%
%

\subsection{Case Visualization of Sequence Representations}
We posit that our STCCR encourages mobility representations from users with different activity topics and intentions to be distinguished. It emphasizes the advantages of the proposed model. Towards that, we compare our pre-trained model with CACSR using t-SNE~\cite{t-sne} to plot the latent space of check-in sequences. Specifically, we select 10 locations with the most truth labels in different categories and their corresponding check-in sequences from each dataset. The learned latent representations of each check-in are projected to the 2D space. Fig.~\ref{case_study} plots the learned representations of check-in sequences on two datasets, where we can observe an apparent clustering effect of sequences. This means that our model can effectively capture the high-level semantic purpose of different users, which is essential for downstream tasks such as next location prediction and user trajectory user link. This result also implies that a better representation with good discriminability is critical for human mobility learning.

\section{Conclusions}
We present STCCR, a Contrastive Spatial-Temporal Cross-view Representation method, to enhance the understanding of human movement patterns in check-in sequences. 
To address the limitations of existing representation learning approaches, STCCR leverages a cross-view contrastive framework that considers both spatial topic and temporal intention views. This innovative approach enables the exploration of macroscopic semantics and associations from different viewpoints, leading to improved representation learning for check-in sequences. 
Furthermore, we introduce an angular momentum contrastive method that effectively captures the inherent uncertainty and temporal intention within the time series data. Through contrastive clustering of spatial topics, we uncover shared spatial activity patterns, enhancing the understanding of human mobility. 
Experimental evaluations on real-world check-in datasets demonstrate the superiority and versatility of our proposed STCCR model.
\ifCLASSOPTIONcompsoc
  \section*{Acknowledgments}
\else
  \section*{Acknowledgment}
\fi
This work was supported by the National Natural Science Foundation of China (No. 62272033) and the China Postdoctoral Science Foundation (Grant No. 2023T160044). 
\ifCLASSOPTIONcaptionsoff
  \newpage
\fi

\bibliographystyle{IEEEtran}
\bibliography{IEEEabrv,main}
\begin{IEEEbiography}[{\includegraphics[width=1in,height=1.25in,clip,keepaspectratio]{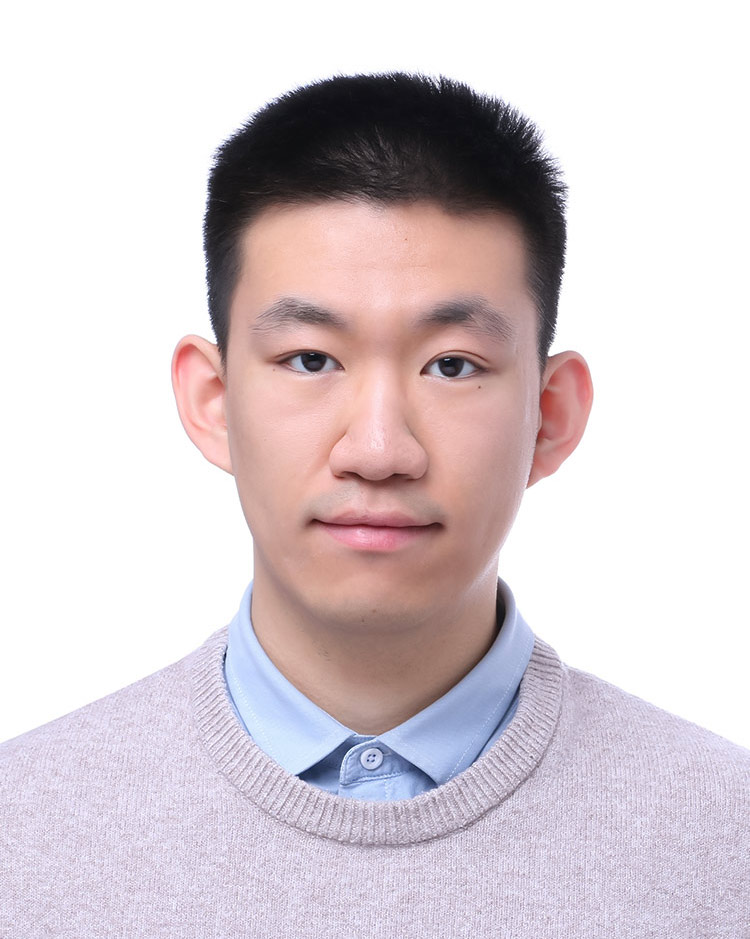}}]{Letian Gong}
received the B.S. degree in computer science from Beijing Jiaotong University, Beijing, China, in 2021.

He is currently working toward the Ph.D. degree in the School of Computer and Information Technology, Beijing Jiaotong University. His interest falls in area of deep learning and data mining, especially their applications in spatial-temporal data mining.
\end{IEEEbiography}

\begin{IEEEbiography}[{\includegraphics[width=1in,height=1.25in,clip,keepaspectratio]{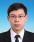}}]{Huaiyu Wan}
received the Ph.D. degree in computer science and technology from Beijing Jiaotong University, Beijing, China, in 2012.

He is a Professor with the School of Computer and Information Technology, Beijing Jiaotong University. His current research interests focus on spatial-temporal data mining, social network mining, and user behavior analysis.
\end{IEEEbiography}

\begin{IEEEbiography}[{\includegraphics[width=1in,height=1.25in,clip,keepaspectratio]{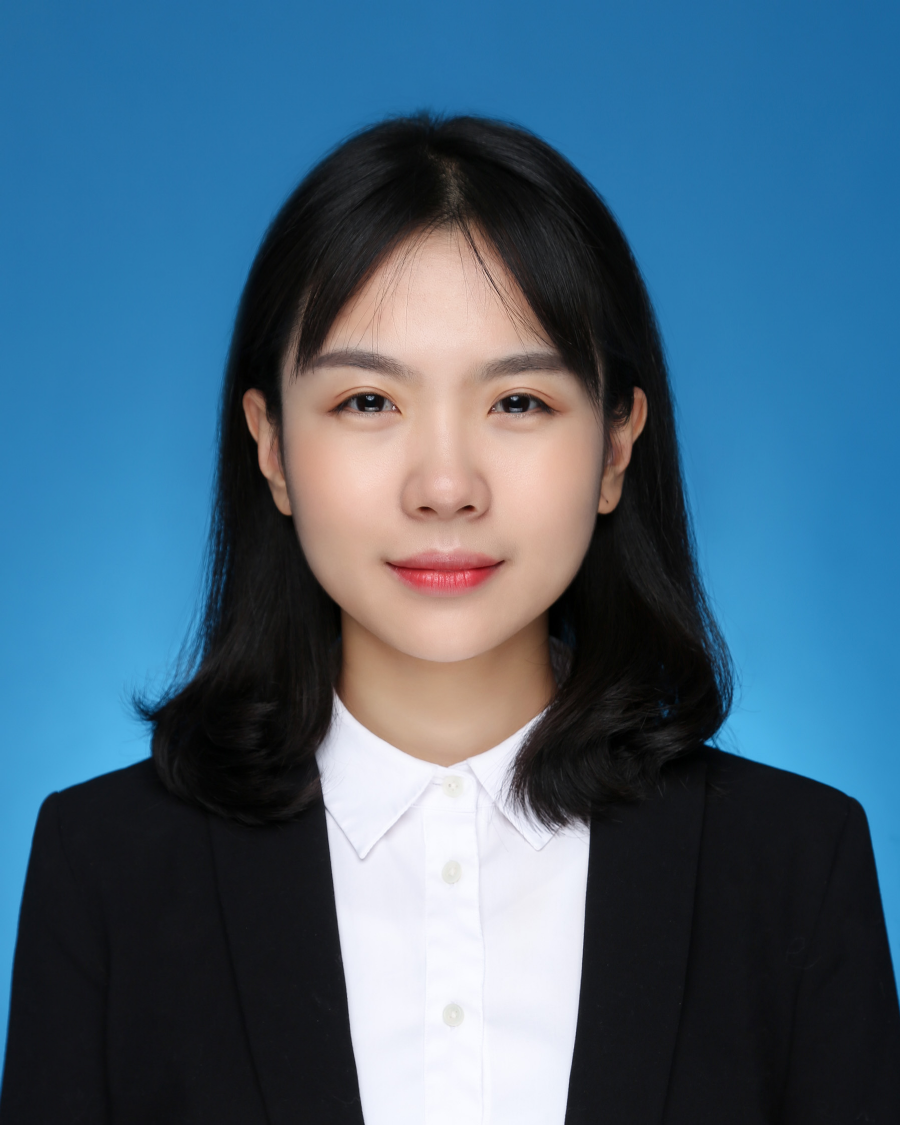}}]{Shengnan Guo} received the Ph.D. degree in computer science from Beijing Jiaotong University, Beijing, China, in 2021. 

She is a lecturer at the School of Computer and Information Technology, Beijing Jiaotong University. Her research interests focus on spatial-temporal data mining and intelligent transportation systems.
\end{IEEEbiography} 

\begin{IEEEbiography}[{\includegraphics[width=1in,height=1.25in,clip,keepaspectratio]{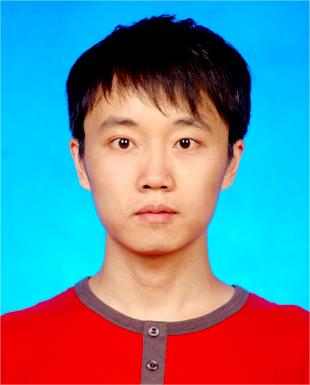}}]{Xiucheng Li}
received the Ph.D. student at the School of Computer Science and Engineering, Nanyang Technological University, Singapore.

At present, he is an assistant professor at the School of Computer Science and Technology, Harbin Institute of Technology (Shenzhen). He is focused on representation learning, generative models and probabilistic inference, spatial and time series data analysis.
\end{IEEEbiography}

\begin{IEEEbiography}[{\includegraphics[width=1in,height=1.25in,clip,keepaspectratio]{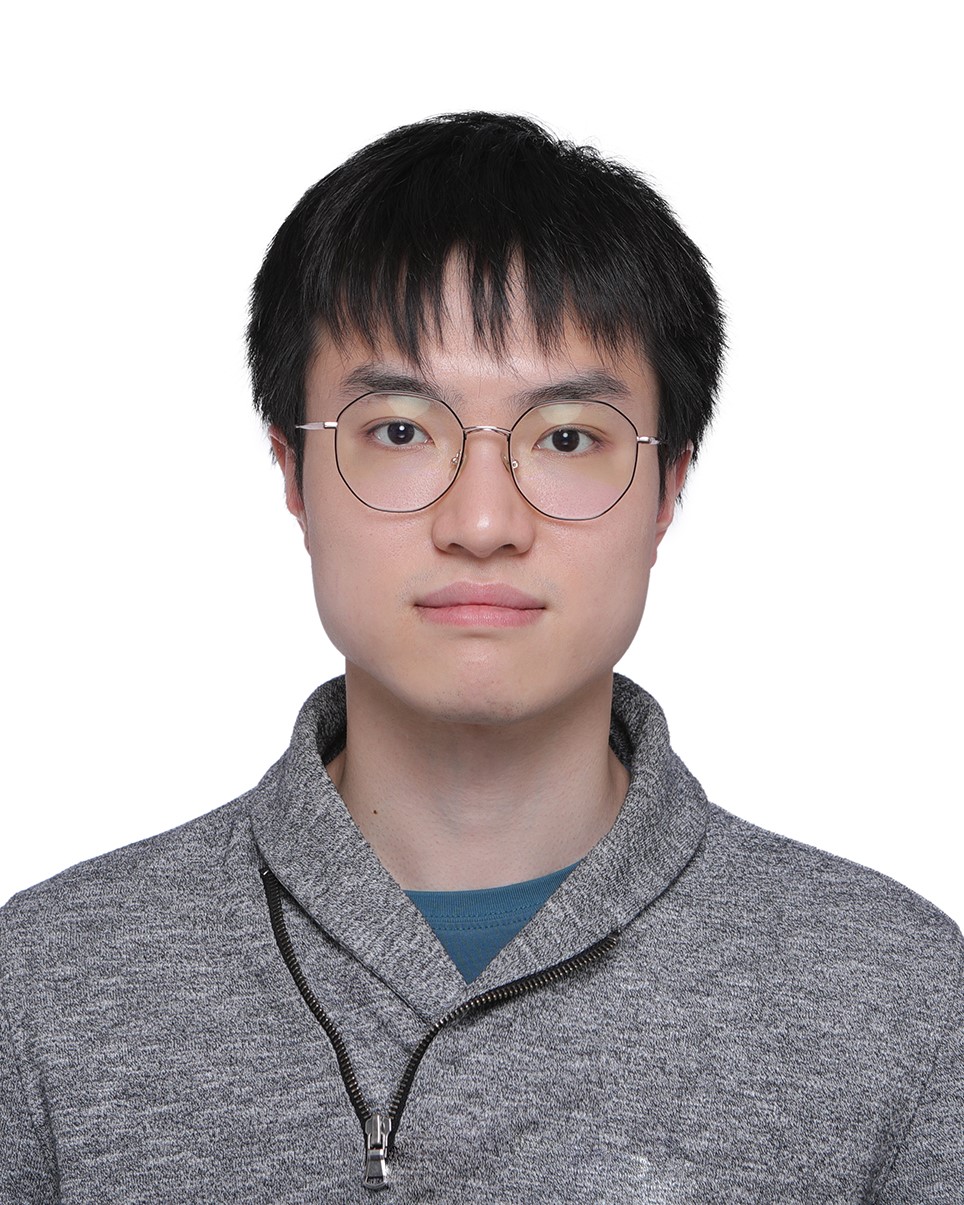}}]{Yan Lin}
  received the B.S. degree in computer science from Beijing Jiaotong University, Beijing, China, in 2019.   
  He is currently working toward the Ph.D. degree from the School of Computer and Information Technology, Beijing Jiaotong University. His research interests include spatial-temporal data mining and graph neural networks.
\end{IEEEbiography}

\begin{IEEEbiography}[{\includegraphics[width=1in,height=1.25in,clip,keepaspectratio]{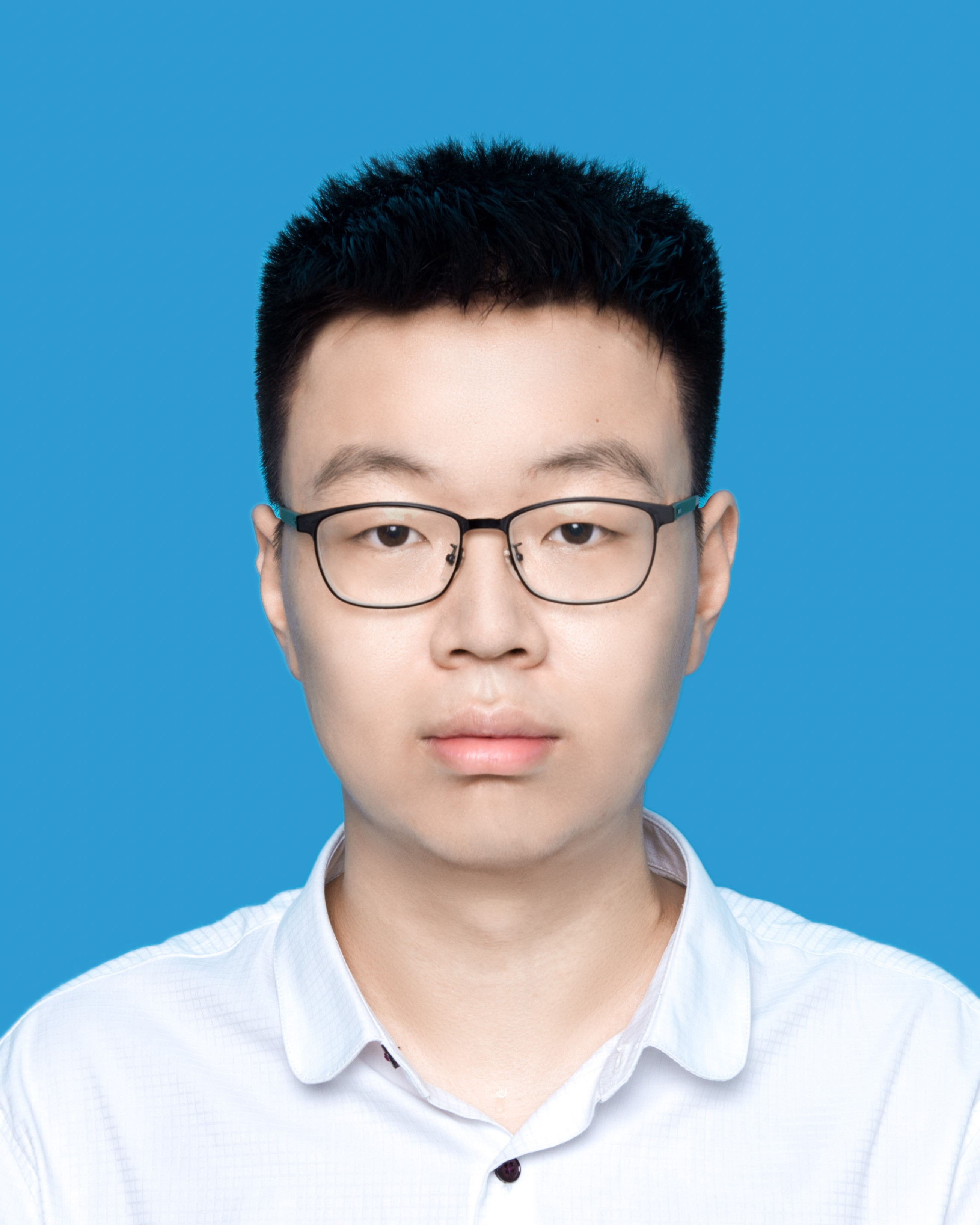}}]{Erwen Zheng}
  received the B.S. degree in mathematics from Beijing Jiaotong University, Beijing, China, in 2023.
 
 He is currently working toward the M.S. degree in the School of Computer and Information technology, Beijing Jiaotong University. His interest falls in area of deep learning and data mining, especially their applications in spatial-temporal data mining.
\end{IEEEbiography}

\begin{IEEEbiography}[{\includegraphics[width=1in,height=1.25in,clip,keepaspectratio]{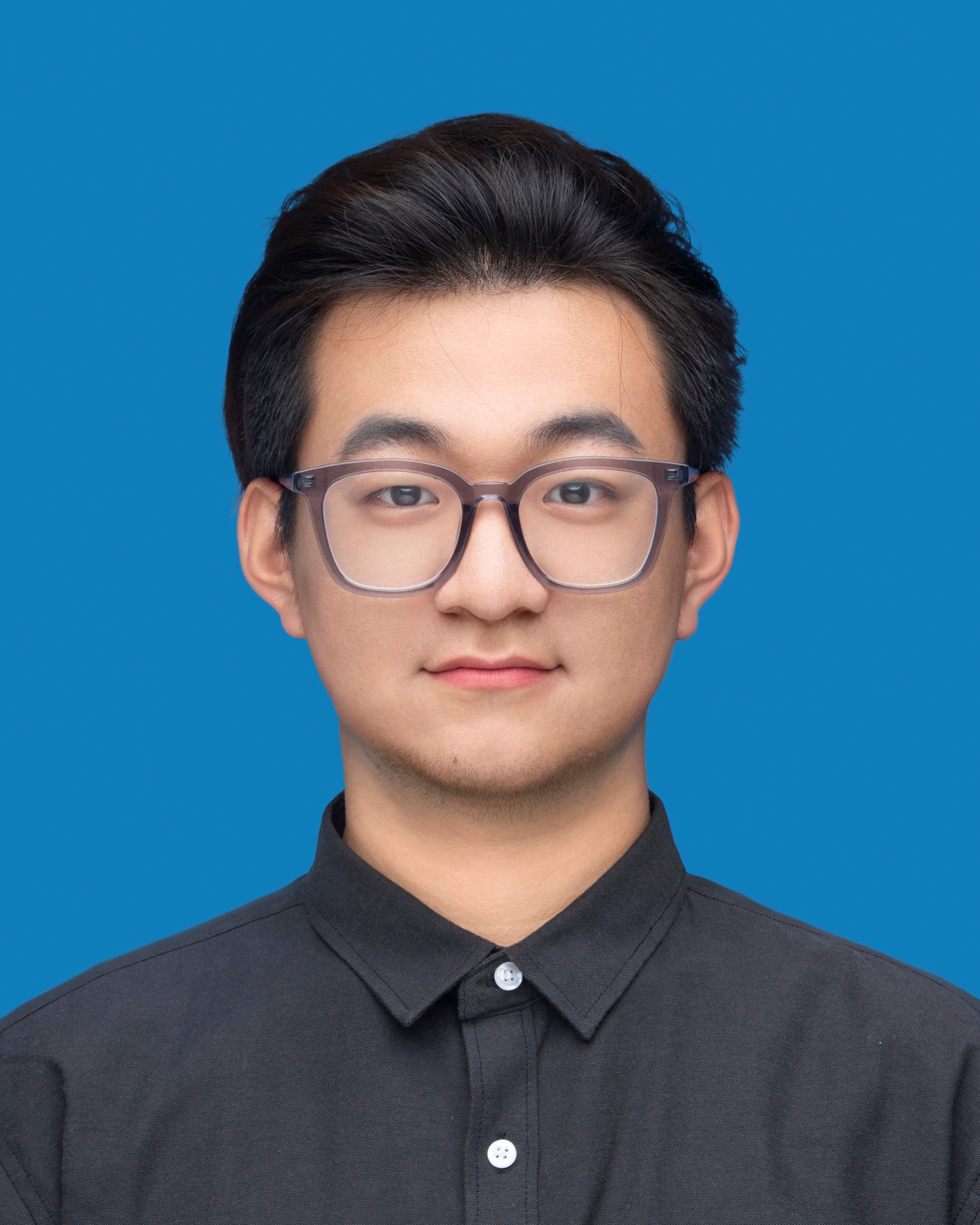}}]{Tianyi Wang}
 received the B.S. degree in computer science from Beijing Jiaotong University, Beijing, China, in 2022.
 
He is currently working toward the M.S. degree in the School of Computer and Information technology, Beijing Jiaotong University. His interest falls in area of deep learning and data mining, especially their applications in spatial-temporal data mining.
\end{IEEEbiography}

\begin{IEEEbiography}[{\includegraphics[width=1in,height=1.25in,clip,keepaspectratio]{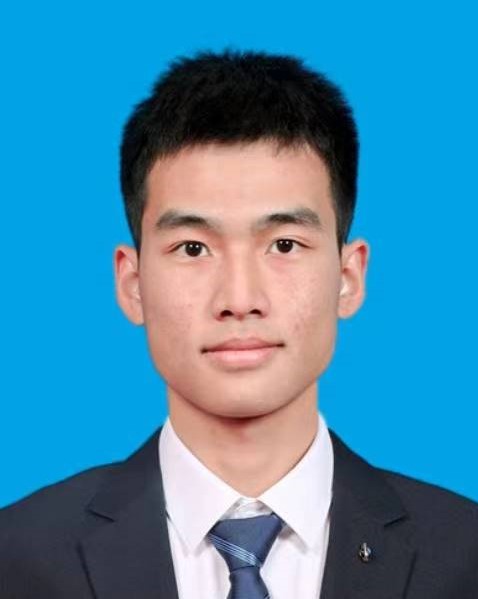}}]{Zeyu Zhou}
 received the B.S. degree in mathematics from Beijing Jiaotong University, Beijing, China, in 2022.
 
He is currently working toward the M.S. degree in the School of Computer and Information technology, Beijing Jiaotong University. His interest falls in area of deep learning and data mining, especially their applications in spatial-temporal data mining.
\end{IEEEbiography}



\begin{IEEEbiography}[{\includegraphics[width=1in,height=1.25in,clip,keepaspectratio]{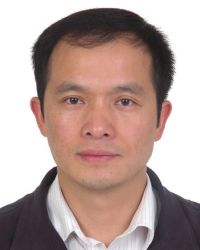}}]{Youfang Lin}
received the Ph.D. degree in signal and information processing  from Beijing Jiaotong University, Beijing, China, in 2003.

He is a Professor with the School of Computer and Information Technology, Beijing Jiaotong University. His main fields of expertise and current research interests include big data technology, intelligent systems, complex networks, and traffic data mining.
\end{IEEEbiography}

\end{document}